\pdfoutput=1
\documentclass[11pt]{article}

\usepackage{ACL2023}
\usepackage{times}
\usepackage{latexsym}
\usepackage{graphicx}
\usepackage{amsmath}
\usepackage{tabularx} % Add this line to your preamble
\usepackage{hyperref}

\usepackage{float}  % Add this to your preamble

\usepackage[T1]{fontenc}
\usepackage[utf8]{inputenc}

\usepackage{microtype}

\usepackage{inconsolata}
\usepackage{placeins} % Include this in your preamble
\usepackage{listings}

\usepackage[inline]{trackchanges}
\addeditor{YJ}

\title{A Comparative Study of Learning Paradigms in Large Language Models via Intrinsic Dimension
}

\author{Saahith Janapati \\
  University of Virginia \\
  \texttt{jax4zk@virginia.edu} \\\And
  Yangfeng Ji \\
  University of Virginia\\
  \texttt{yj3fs@virginia.edu} \\}

\begin{document}
\maketitle
\begin{abstract}
The performance of Large Language Models (LLMs) on natural language tasks can be improved through both supervised fine-tuning (SFT) and in-context learning (ICL), which operate via distinct mechanisms. SFT updates the model’s weights by minimizing loss on training data, whereas ICL leverages task demonstrations embedded in the prompt, without changing the model’s parameters. This study investigates the effects of these learning paradigms on the hidden representations of LLMs using Intrinsic Dimension (ID). We use ID to estimate the number of degrees of freedom between representations extracted from LLMs as they perform specific natural language tasks. We first explore how the ID of LLM representations evolves during SFT and how it varies due to the number of demonstrations in ICL. We then compare the IDs induced by SFT and ICL and find that ICL consistently induces a higher ID compared to SFT, suggesting that representations generated during ICL reside in higher dimensional manifolds in the embedding space. \footnote{Code is available at the following \href{https://github.com/saahithjanapati/intrinsic-dimension-of-learning-paradigms}{GitHub repo}.}

\end{abstract}

\section{Introduction}

Large Language Models (LLMs) have transformed the field of Natural Language Processing through their general natural language understanding capabilities, which can be applied to a broad range of tasks. The performance of an LLM on a specific task can be improved through two primary learning paradigms: supervised fine-tuning (SFT) and in-context learning (ICL). SFT adapts pre-trained models to specific tasks by updating their parameters, while ICL requires no parameter updates, relying instead on task-specific demonstrations within the model’s context window. Despite their widespread success, how these methods influence a model’s internal representation space is still not fully understood.

Intrinsic dimension (ID) is a useful metric for assessing the geometric complexity of a model’s representations. It quantifies the number of degrees of freedom in the representation space, serving as a measure of the complexity of the underlying manifolds where the embeddings reside.

In this work, we analyze the intrinsic dimension (ID) of hidden representations across model layers during task execution under both supervised fine-tuning (SFT) and in-context learning (ICL). Specifically, we explore:

\begin{itemize}
    \item How fine-tuning duration influences ID of representations on both training and validation data.
    \item How the number of demonstrations used in ICL affects ID of representations.
\end{itemize}

Our findings reveal that (1) the ID sometimes decreases during the early stages of fine-tuning but generally increases in the later stages, and (2) the ID increases initially with more demonstrations in ICL, then either plateaus or decreases as the number of demonstrations continues to rise.

We then conduct experiments directly comparing the intrinsic dimensions of ICL and fine-tuning across several models and datasets. We find that the intrinsic dimensions of representations from fine-tuned models are generally lower than those from models using ICL, even though the fine-tuned models achieve higher accuracy than the ICL models. Additionally, our results suggest that ID may serve as a practical heuristic for selecting the optimal number of demonstrations in ICL to maximize performance while minimizing input length. These findings shed light on the differing impacts that the two learning paradigms have on the representation space of LLMs.

\begin{figure*}[t]
    \centering
    \includegraphics[width=\textwidth]{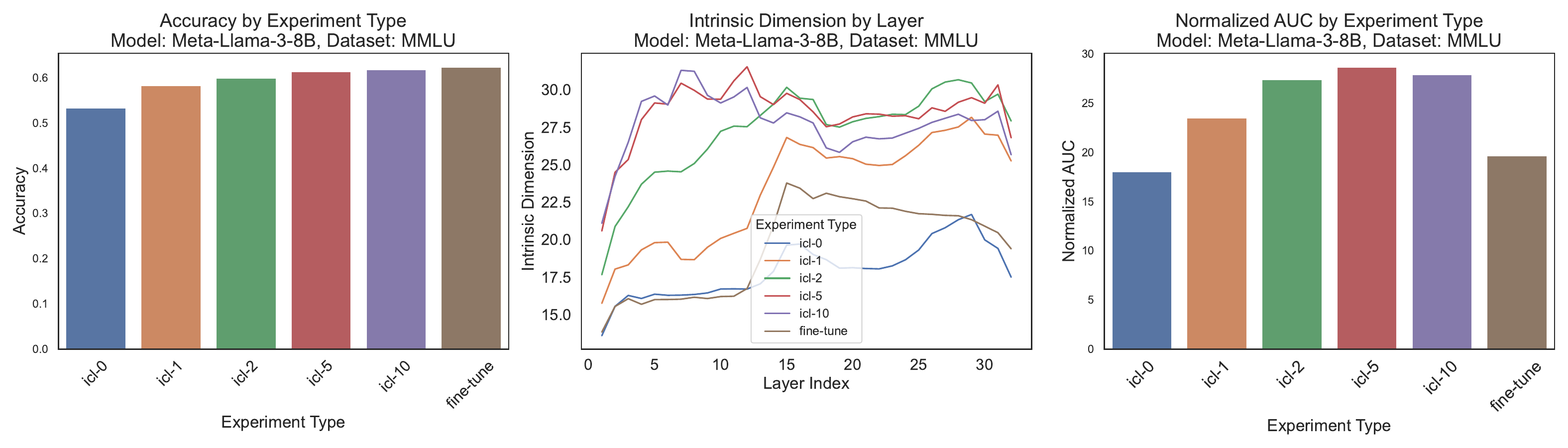}
    \caption{Accuracy, intrinsic dimension, and normalized AUC for the Llama-3-8B model on the MMLU dataset. (a) Fine-tuning achieves the highest accuracy. (b) ICL produces intermediate representations with higher intrinsic dimensions across model layers compared to zero-shot (ICL-0) and fine-tuned models. (c) Normalized AUC increases with the number of demonstrations in ICL, while fine-tuned models exhibit lower AUC.}
    \label{fig:compare_both}
\end{figure*}

\section{Background}

\subsection{Decoder Transformer Architecture}
\label{sec:decoder-transformer-arch}
LLMs are built on the Transformer decoder architecture, which processes token sequences through a series of Transformer layers. In each layer, token representations are updated via self-attention that considers only the preceding tokens from the previous layer, progressively encoding information for the next-token prediction task. The final layer then uses the representation of the last token to predict the next token in the sequence. In this work, we analyze the intrinsic dimension of the representations corresponding to the last token of sequences where LLMs are prompted to perform specific natural language tasks.

\subsection{Intrinsic Dimension Estimation}
Intrinsic dimension (ID) refers to the minimal number of variables required to capture the essential structure of high-dimensional data. Although modern neural networks operate in high-dimensional spaces (e.g., the hidden representations of Llama-3-8B span 4096 dimensions), the representations corresponding to a specific dataset or task often lie on a manifold of much lower dimension. This occurs because the network disentangles and extracts the most relevant lower-dimensional features needed to complete the task.

According to the manifold hypothesis, real-world data typically resides on a low-dimensional manifold \citep{goodfellow2016deep}. Therefore, to effectively solve tasks—such as next-token prediction—neural networks must learn representations that align with this low-dimensional structure. Consequently, the intrinsic dimension of data representations provides unique insight into the complexity of the representation spaces constructed across the layers of a neural network.

In this work, we estimate the intrinsic dimension (ID) of our representations using the \textbf{TwoNN estimator}, as introduced by \citet{facco2017estimating}. We chose this method because of its simplicity, computational efficiency, and robustness when handling datasets with non-uniform densities and high-dimensional curvature---common challenges in neural network representations.

The TwoNN estimator operates on a set of points by computing the distances to each point’s first (\(r_1\)) and second (\(r_2\)) nearest neighbors. For a given point \(x\), the ratio 
\[
\mu = \frac{r_2}{r_1}
\]
is calculated. The intrinsic dimension \(d\) is then derived from the empirical cumulative distribution function (CDF) of \(\mu\). Specifically, the log-linear relationship between \(\log(\mu)\) and \(\log\bigl(1 - F_{\text{emp}}(\mu)\bigr)\), where \(F_{\text{emp}}(\mu)\) is the empirical CDF, is used to estimate \(d\):
\[
d = -\frac{\log\bigl(1 - F_{\text{emp}}(\mu)\bigr)}{\log(\mu)}
\]

The TwoNN estimator has been successfully applied in several prior works analyzing the intrinsic dimension of neural network representations, including \citet{sharma2022scaling}, \citet{ansuini2019intrinsic}, \citet{valeriani2024geometry}, and specifically in large language models (LLMs) by \citet{cheng2023bridging} and \citet{lee2024geometric}. We also validate the correlation between the TwoNN estimator and another widely used intrinsic dimension estimator---the Maximum Likelihood Estimator introduced by \citet{levina2004maximum}---in Appendix~\ref{sec: validate-twonn} as a sanity check.

\section{Related Works}

\subsection{Supervised Fine-Tuning in LLMs}
Pre-trained LLMs can be quickly adapted to improve performance on natural language tasks through supervised fine-tuning, which updates the model’s parameters via gradient descent on task-specific training examples.

\citet{aghajanyan2020intrinsic} show that fine-tuning large language models often requires updating only a low-dimensional subspace of parameters to achieve near-optimal performance. (Note that their work focuses on the intrinsic dimension of the parameter space, whereas our work examines the intrinsic dimension of the representation space.) Building on this, \citet{hu2021lora} introduce Low-Rank Adaptation (LoRA), a method that injects low-rank matrices into the weight matrices for fine-tuning instead of updating all parameters. We employ LoRA for all our fine-tuning experiments.

\subsection{In-Context Learning}
Introduced in GPT-3 by \citet{brown2020language}, ICL (or few-shot learning) refers to the ability of LLMs to learn to perform a task in a single forward pass, using (input, output) pairs embedded in a prompt.

\citet{dai2022can} provides evidence that ICL operates as implicit meta-optimization, where GPT models perform a gradient-like update via attention mechanisms during the forward pass. This suggests that ICL replicates fine-tuning behavior; specifically, they demonstrate that attention outputs and weights are updated in a direction similar to that of fine-tuning.

\citet{xie2021explanation} explain in-context learning as implicit Bayesian inference, where large language models infer latent document-level concepts during pretraining. These inferred concepts are then leveraged at test time to solve tasks based on the input-output examples provided in prompts.

Expanding the ICL paradigm to long-context models, \citet{agarwal2024many} studied many-shot ICL, in which hundreds or thousands of task examples are used to improve the performance of frontier models. Their work finds that an increasing number of demonstrations generally improves model performance on a variety of complex tasks, such as mathematical problem-solving.

\subsection{Intrinsic Dimension in Deep Learning}

\citet{ansuini2019intrinsic} investigated the intrinsic dimensionality (ID) of data representations across various convolutional neural networks (CNNs) for image classification. They observed a consistent “hunchback” pattern in ID evolution—an initial increase in the early layers followed by a progressive decrease in later layers.

\citet{valeriani2024geometry} extended this analysis to protein language models and image transformers, finding that the evolution of representations across layers of these models is also marked by distinct phases of ID growth and compression.

\citet{yin2024characterizing} explore the use of Local Intrinsic Dimension (LID) to detect untruthful outputs from LLMs. Their study reveals that truthful outputs typically exhibit lower LIDs compared to hallucinated ones, suggesting that LID can serve as a signal for truthfulness in LLM generations. They also identify a positive relationship between the ID of data representations and validation performance during fine-tuning.

\citet{cheng2023bridging} demonstrate that intrinsic dimension correlates with fine-tuning ease and perplexity, with low-dimensional representations enabling faster task adaptation. Moreover, they find that ID values are consistent across model sizes, supporting the manifold hypothesis and suggesting that LLMs trained on similar data recover comparable intrinsic dimensions.

Of particular relevance to our study is the concurrent work of \citet{doimo2024representation}, which examines the internal representations of LLMs solving tasks from the MMLU dataset using both ICL and SFT. Their analysis reveals that ICL forms semantic clusters in the early layers, while SFT sharpens these clusters in later layers for task-specific answers. Moreover, they observe that intrinsic dimension (ID) increases with a higher number of demonstrations in ICL, and that SFT generally induces a higher ID compared to ICL. In contrast, our findings indicate that beyond a certain range of ICL demonstrations, ID may plateau or even decrease, and that SFT consistently induces a lower ID than ICL.

To our knowledge, our work is the first to systematically analyze and compare intrinsic dimension across the two learning paradigms for numerous datasets and models. We further provide in-depth analyses of how ID is affected by various factors within each paradigm, such as the number of gradient steps in SFT and the number of demonstrations in ICL.

\subsection{Intrinsic Dimension and Neural Network Scaling Laws}
\citet{sharma2022scaling} propose that the power-law scaling of neural network performance is rooted in the intrinsic dimensionality (ID) of the data manifold. They empirically demonstrate that the ID of learned representations, particularly in the final hidden layer, directly relates to the scaling exponent. Their theory, predicting a scaling exponent of approximately $\alpha \approx 4/d$ (where $d$ is ID), suggests that neural networks achieve efficient scaling by effectively performing regression on a lower-dimensional data manifold, thus linking model capacity to the data's inherent complexity.

\section{Methods}
\label{sec:methods}
We perform experiments using subsets from the following datasets: AG News \citep{zhang2015character}, CoLA \citep{warstadt2018neural}, CommonsenseQA \citep{talmor2018commonsenseqa}, MMLU \citep{hendrycks2020measuring}, MultiNLI \citep{williams2017broad}, QNLI \citep{wang2018glue}, QQP \citep{wang2017bilateral}, and SST2 \citep{socher2013recursive}. 

For these experiments, we utilize the following open-source LLMs: Llama-3-8B \citep{dubey2024llama}, Llama-2-13b, Llama-2-7b \citep{touvron2023llama}, and Mistral-7B-v0.3 \citep{jiang2023mistral}, running them on 6 NVIDIA A6000s. 

For each dataset, we created a training set of 1000 examples and a validation set of 5000 examples. We use the 5000 validation examples to ensure stability of the TwoNN estimator. Details regarding dataset creation can be found in Appendix \ref{sec: dataset-curation-details}. Details of split generations and prompt templates are provided in Appendix \ref{sec: dataset-curation-details}. 

We calculate the accuracy of model responses using the logit probabilities assigned to the tokens corresponding to the possible answers for each question. We mark a response as correct if the probability corresponding to the first token of the correct answer label is the highest.

\subsection{Computing Intrinsic Dimension}
In both Supervised Fine-Tuning (SFT) and In-Context Learning (ICL) paradigms, a language model receives an input sequence of tokens and is tasked with generating an output sequence that answers the given prompt. To quantify the intrinsic dimensionality (ID) of a model's representations for a given dataset, we extract the hidden state activations at each layer of the LLM. Specifically, we focus on the activations corresponding to the \textbf{last token of each input sequence} in the dataset. For a model with $L$ layers and a dataset containing $N$ input sequences, this process yields $L$ sets of hidden state representations. Each set corresponds to a specific layer and comprises $N$ representation vectors (one for each input sequence in the dataset). Subsequently, we compute the intrinsic dimension (ID) for each of these $L$ sets of $N$ vectors. This provides us with an ID estimate for the representation space at each layer. By plotting the Layer Index against the corresponding ID estimates, we construct what we term the \textbf{Intrinsic Dimension Curve}.

\medskip
\noindent To derive a single, aggregated metric that encapsulates the intrinsic dimensionality across all layers of a model, we calculate the \textbf{Normalized Area Under the Curve (AUC)} of the Intrinsic Dimension Curve, defined as follows:
\begin{multline*}
\text{Normalized AUC} = \frac{1}{L}\sum_{i=1}^{L-1} \frac{1}{2}
\left( \text{ID}_i + \text{ID}_{i+1} \right)
\end{multline*}
In this equation, \( \text{ID}_i \) denotes the intrinsic dimension estimate at layer \( i \). The formula employs the trapezoidal rule for numerical integration to approximate the area beneath the Intrinsic Dimension Curve. The normalization by $L$ (the number of layers) enables fair comparisons of intrinsic dimensionality across models with varying depths.

\section{Dynamics of ID during Supervised Fine-Tuning}
\label{sec:result-sft}

\subsection{Supervised Fine-Tuning Experimental Setup}
To investigate the impact of supervised fine-tuning at a granular level, we conduct experiments using the 8 datasets discussed in Section \ref{sec:methods} and the Llama-3-8B and Llama-2-13B models.

Using the training split for each of the datasets, we perform LoRA fine-tuning on the query, key, value, and output projection matrices of attention heads across all layers of the model. For all models, we fine-tune with a batch size of 16 for 15 epochs. For all fine-tuning runs, we use LoRA hyperparameters of $r=64$, $\text{lora\_alpha}=16$, $\text{lora\_dropout}=0.1$, no LoRA bias, and a learning rate of $1e^{-4}$.

During the fine-tuning process for a specific model and dataset, we save a checkpoint of the model after every epoch (\textasciitilde 62 gradient update steps). For each checkpoint, we evaluate the model's accuracy and measure the intrinsic dimension (ID) of the hidden representations on prompts from the training and validation splits for the dataset.

\subsection{Intrinsic Dimension Generally Increases Through Fine-Tuning}
\begin{figure}[t]  % [t] for top placement, can use [h] for here or [b] for bottom
    \centering
    \includegraphics[width=\linewidth]{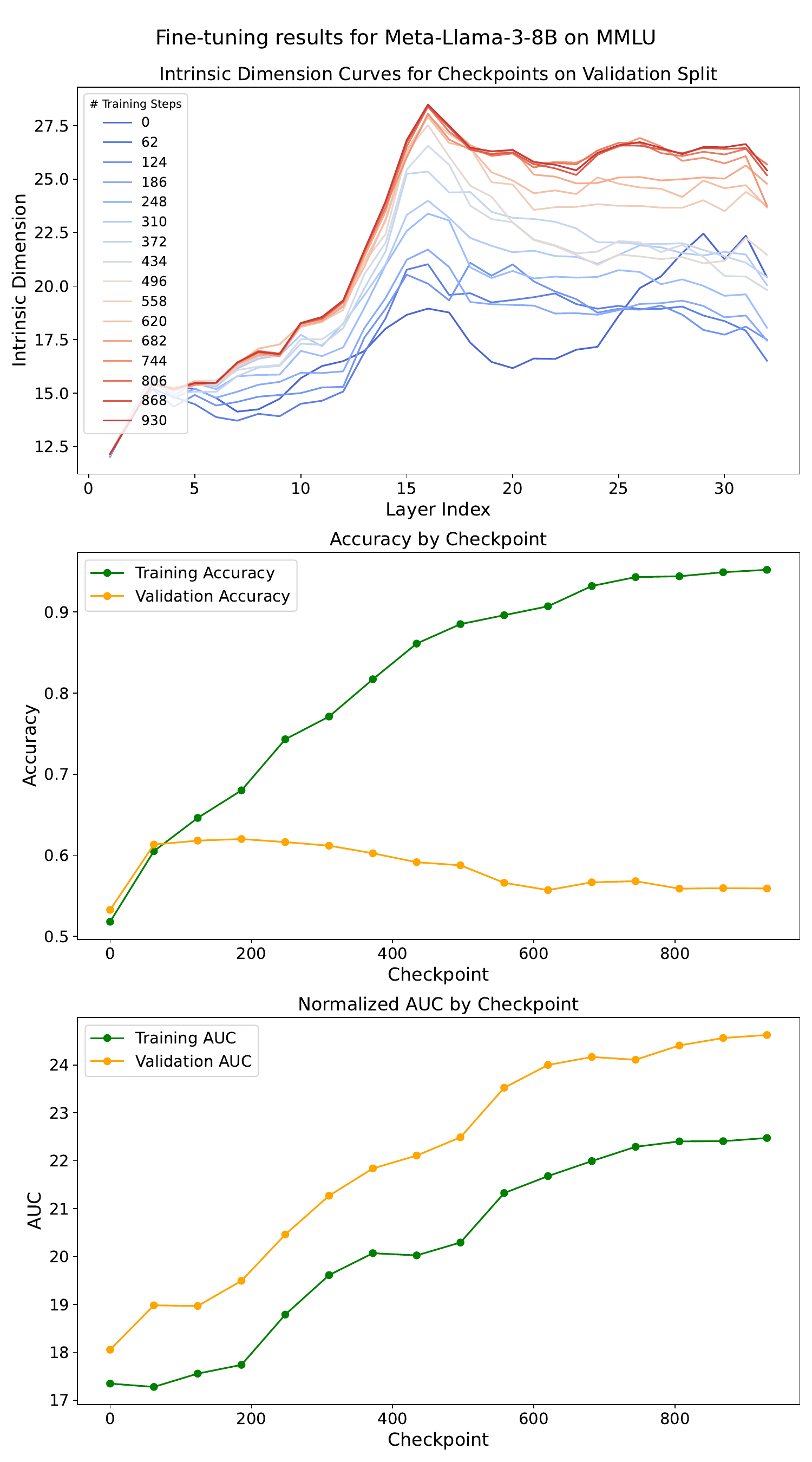}
    \caption{Fine-tuning results for Meta-Llama-3-8B on the MMLU dataset. (a) Intrinsic Dimension curves on the validation split increase across training steps. (b) Training accuracy improves steadily, while validation accuracy plateaus. (c) Normalized AUC for training and validation sets increases throughout fine-tuning.}
    \label{fig:ft-if-mmlu}
\end{figure}

As depicted in Figure \ref{fig:ft-if-mmlu}c, we find that ID of representations corresponding to both training data and validation data sometimes decreases during the initial stages of fine-tuning, but then generally increases as fine-tuning progresses.

We also observe larger changes in ID values for later layers of the models, despite LoRA adaptation being applied on all the layers with the same configuration (Figure \ref{fig:ft-if-mmlu}a).

Additionally, we find that the AUC values of the model on the training set and validation set are often highly correlated with each other during the training process (Figure \ref{fig:ft-if-mmlu}c). Experimental results for all models and datasets can be found in Appendix \ref{sec:detailed-sft}.

Prior work by \citet{yin2024characterizing} found that on Question-Answering datasets, intrinsic dimension of representations is correlated with validation performance and can therefore be used as a heuristic to select final checkpoints. In general, we do not find this trend to hold on the datasets and models we tested. In fact, as shown in Figure \ref{fig:icl-llama-8b-mmlu}, large increases in validation accuracy sometimes coincide with drops in ID on both the training and validation datasets.

\section{Relationship of ID in ICL with Different Numbers of Shots}
\label{sec:icl-results}

\begin{figure*}[t]
    \centering
    \includegraphics[width=\textwidth]{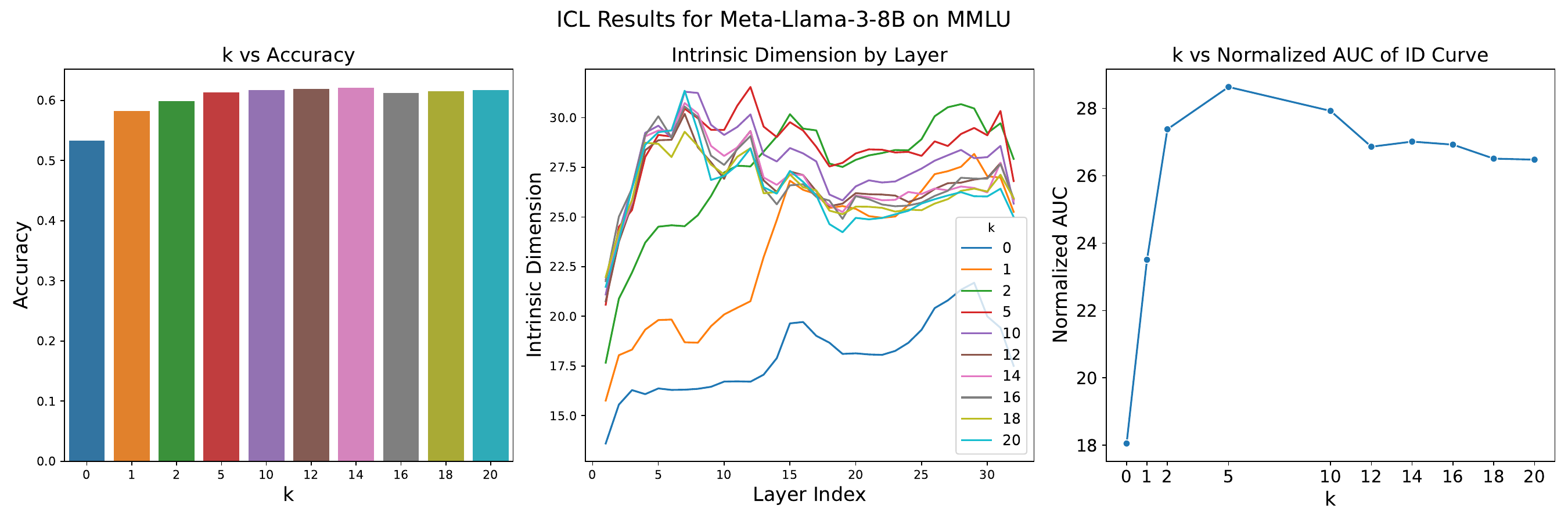}
    \caption{(ICL) results for Meta-Llama-3-8B model on MMLU dataset. (a) Accuracy increases, then plateaus as number of demonstrations increases (b) Intrinsic Dimension (ID) curves for different values of k. (c) Normalized AUC of the ID curves peaks at k=5, which also aligns with saturation of accuracy.}
    \label{fig:icl_single}
\end{figure*}

\subsection{In-Context Learning Experimental Setup}
To investigate the impact of ICL on the ID of model representations, we conduct experiments using the Llama-3-8B and Llama-2-13B models. The datasets included in our evaluation are CommonsenseQA, MMLU, and QNLI.

We evaluate ICL performance using various values of $k$, where $k$ denotes the number of demonstrations in the ICL prompt. The values considered are $k \in \{0, 1, 2, 5, 10, 12, 14, 16, 18, 20\}$. Note that $k=0$ serves as a baseline, representing the model's performance in the absence of both ICL and SFT.

For each $k$ and dataset, we generate 5000 ICL prompts (one for each element of the validation split of the dataset). Each ICL prompt includes $k$ unique demonstrations, or (input, output) pairs, randomly sampled from the training set. While we ensure that demonstrations within a single prompt are unique, they may be reused across different prompts.

\subsection{ID Has a Non-Linear Relationship with Number of Demonstrations}
We observe that ID values across layers can fluctuate until a threshold value of \( k \) (typically around 5 to 10 for most model configurations), after which they either plateau or steadily decrease for larger values of \( k \) (see Figure \ref{fig:icl_single}c). Results for all model and dataset configurations are provided in Appendix \ref{sec:detailed-icl}. This observation extends the findings of \citet{doimo2024representation}, who found that ID increased as \( k \) was varied from 0, 1, 2, and 5, by demonstrating that beyond a certain number of demonstrations, the trend can reverse.

We observe that across most (model, dataset) combinations, the shapes of the intrinsic dimension (ID) curves correlate strongly with each other for \( k \geq 2 \).

Due to our procedure of selecting demonstrations with replacement, we suspected that the plateau in ID for larger values of \( k \) might be caused by a greater number of demonstrations shared across prompts. We hypothesized that shared demonstrations could make representations corresponding to these prompts artificially similar, thereby skewing ID results. To test this, we performed additional experiments using a larger number of dataset elements from the CommonsenseQA, QNLI, and AG News datasets, which contain enough training elements to ensure that demonstrations are not reused in prompts for more than one element of the validation set. We observed the same trend—an increase followed by a general plateau in the ID—suggesting that the plateau is likely not due to the reuse of demonstrations among the prompts. Full results for this experiment can be found in Appendix \ref{sec: de-duplication}.

Furthermore, we find that peaks in the \( k \) versus AUC relationship align with peak (or near-peak) accuracy in 5 out of the 6 ICL experiments we conducted. Thus, the \( k \) value corresponding to the peak ID may serve as a practical indicator of the optimal number of demonstrations to use for ICL, maximizing performance while minimizing input length.

One hypothesis for why ID plateaus or slightly decreases as \( k \) increases is that more demonstrations allow the model to more effectively capture the underlying task conveyed by the demonstrations, causing representations corresponding to different inputs to become more similar. This idea is supported by previous theoretical analysis of ICL by \citet{xie2021explanation}, which posits that a greater number of demonstrations helps the model more effectively infer the latent concept across demonstrations.

Finally, we find that across most experiments, accuracy either steadily increases or plateaus with higher numbers of demonstrations (Figure \ref{fig:icl_single}a).

\section{Comparing Intrinsic Dimension of In-Context Learning and Supervised Fine-Tuning}

\label{sec:acc-table-box}
\begin{table*}[ht]
\centering
\begin{tabular}{lcccccc}
\hline
\textbf{Dataset} & \textbf{ICL-0} & \textbf{ICL-1} & \textbf{ICL-2} & \textbf{ICL-5} & \textbf{ICL-10} & \textbf{Finetune 1K}\\
\hline
\textbf{SST-2} & 0.685 & 0.633 & 0.731 & 0.807 & 0.832 & \textbf{0.944} \\
\textbf{CoLA} & 0.720 & 0.723 & 0.735 & 0.746 & 0.742 & \textbf{0.750} \\
\textbf{QNLI} & 0.517 & 0.513 & 0.555 & 0.590 & 0.585 & \textbf{0.761} \\
\textbf{QQP} & 0.417 & 0.462 & 0.485 & 0.508 & 0.519 & \textbf{0.707} \\
\textbf{MNLI} & 0.374 & 0.367 & 0.387 & 0.414 & 0.431 & \textbf{0.676} \\
\textbf{AGNews} & 0.638 & 0.573 & 0.712 & 0.772 & 0.809 & \textbf{0.881} \\
\textbf{CommonsenseQA} & 0.199 & 0.375 & 0.417 & 0.470 & 0.492 & \textbf{0.500} \\
\textbf{MMLU} & 0.449 & 0.488 & 0.511 & 0.524 & 0.531 & \textbf{0.542} \\
\hline
\end{tabular}
\caption{Average accuracy results for Datasets across ICL and SFT settings. SFT obtains the highest average accuracy for all datasets. Accuracy increases and then plateaus for higher number of demonstrations.}
\label{tab:acc-results}
\end{table*}

\subsection{Experiment Setup for Comparative Analysis}
We conduct a series of experiments to directly compare the ID curves obtained from both SFT and ICL, following similar setups as discussed in Sections \ref{sec:result-sft} and \ref{sec:icl-results}. For the fine-tuning experiments in this section, we train for only 4 epochs and measure the accuracy and ID solely at the final checkpoint. This choice is motivated by the observation in Section \ref{sec:result-sft} that models tend to overfit beyond 4 epochs across the tested datasets.

For the ICL experiments, we consider values of \(k \in \{0, 1, 2, 5, 10\}\) for the number of demonstrations. These values are popular in practice, and our previous experiments in Section \ref{sec:icl-results} indicate that ID curves tend to plateau when \(k \geq 10\). We perform these experiments on all 8 datasets and 4 models discussed in Section \ref{sec:methods}.

\subsection{In-context Learning Induces Higher IDs Compared to Fine-Tuning}
We find that across all datasets and models, ICL prompts with \( k \geq 5 \) consistently induces higher intrinsic dimensions (IDs) across all layers compared to both SFT and 0-shot prompts (see Figures \ref{fig:compare_both}b and \ref{fig:compare_both}c). This contrasts with the findings of \citet{doimo2024representation}, who find that SFT models often induces higher ID than models performing ICL.

\begin{figure}[t]  % [t] for top placement, can use [h] for here or [b] for bottom
    \centering
    \includegraphics[width=\linewidth]{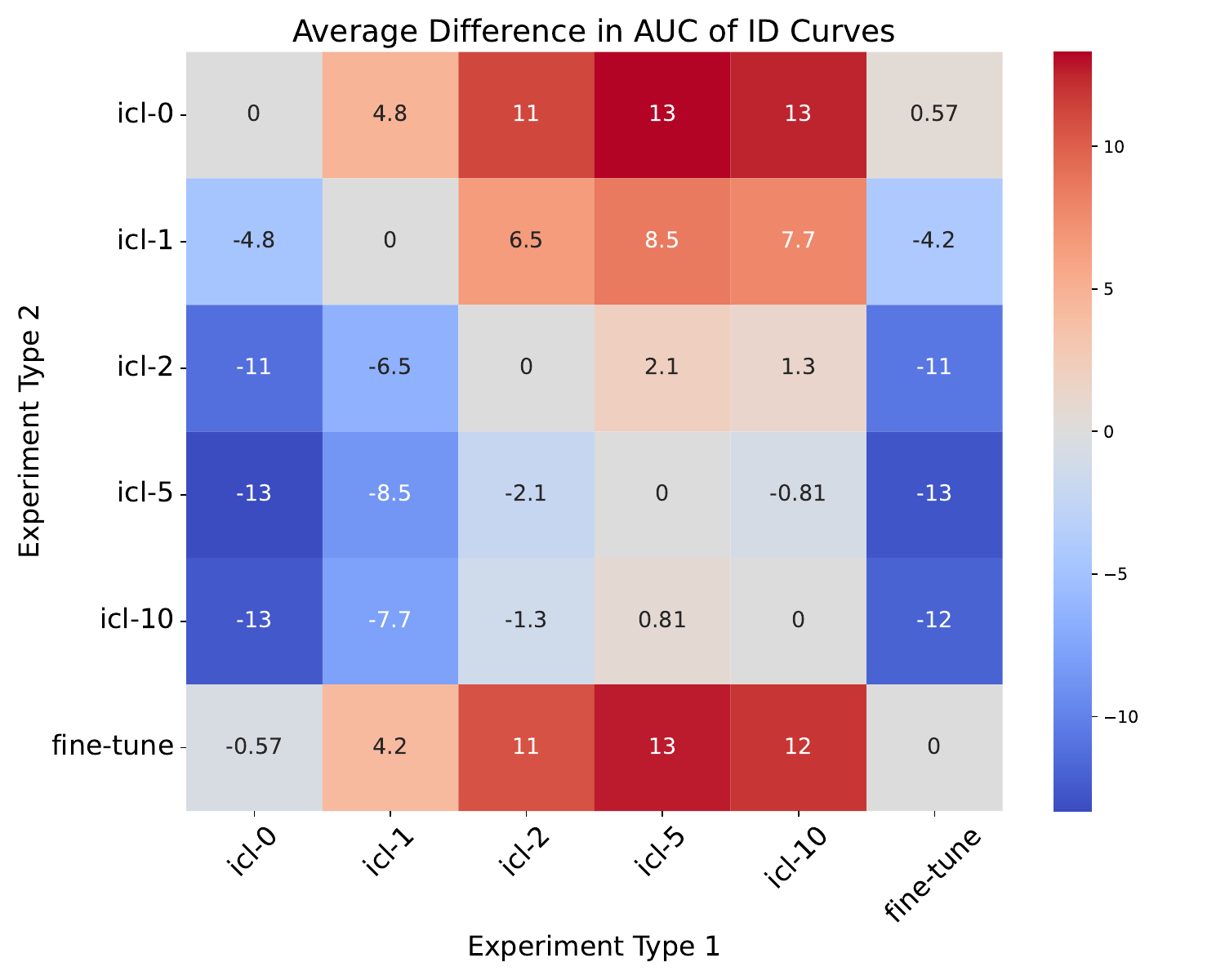}
    \caption{Heatmap showing the average differences in normalized AUC of ID curves between pairs of learning paradigms. Each value represents the average difference (Experiment Type 1 - Experiment Type 2), computed across all (model, dataset) pairs.}
    \label{fig:heatmap}
\end{figure}

\begin{figure}[t]  % [t] for top placement, can use [h] for here or [b] for bottom
    \centering
    \includegraphics[width=\linewidth]{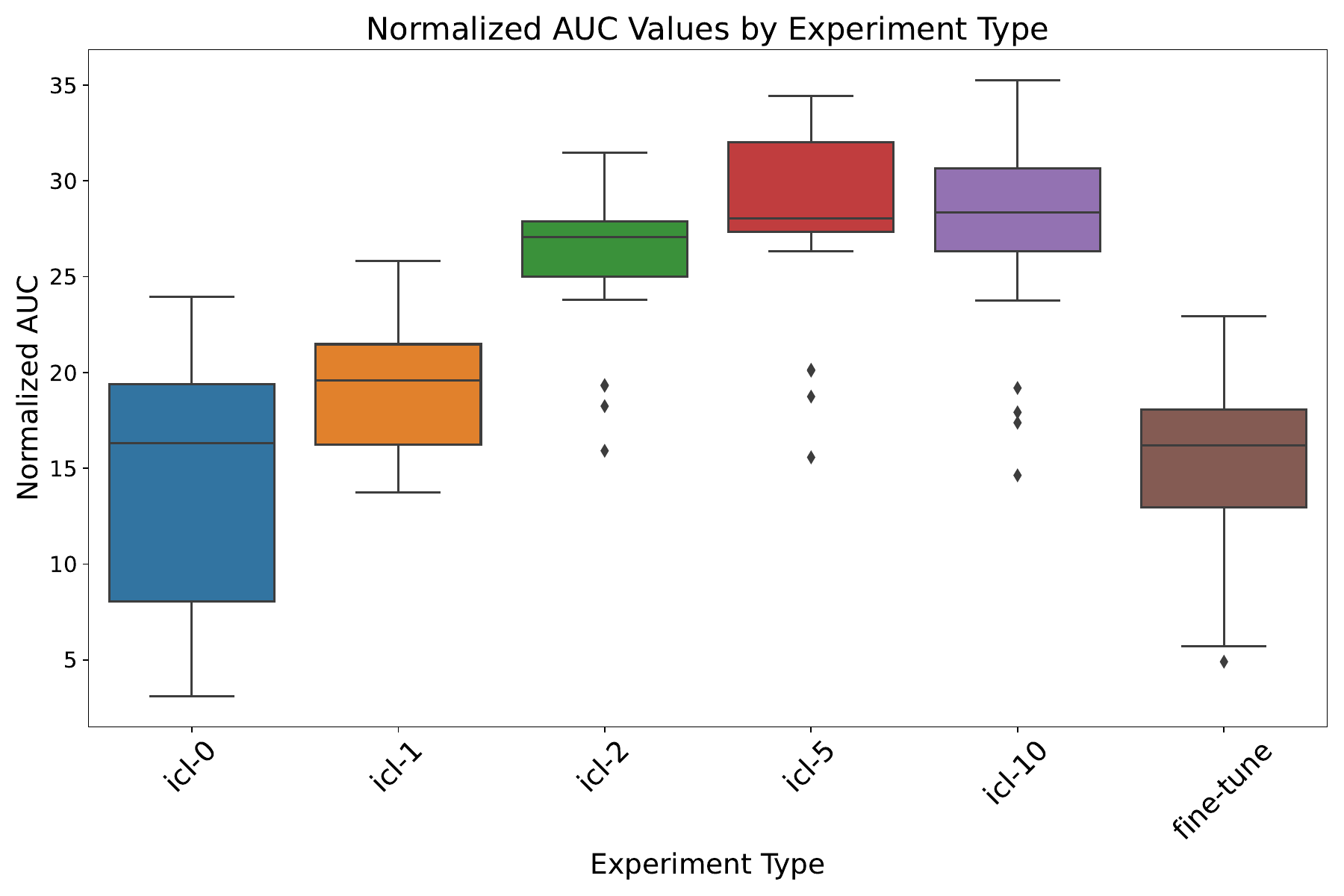}
    \caption{Boxplot displaying the distribution of normalized AUC values for different learning paradigms. Each point corresponds to the normalized AUC value for a (model, dataset) pair. The median normalized AUC peaks with 5-shot ICL, while values for SFT are closer to the 0-shot baseline (icl-0).}
    \label{fig:auc-learning-paradigm}
\end{figure}

We also find that the ID values of models fine-tuned with 1000 samples tend to remain similar to the original ID of the baseline model on a zero-shot prompt (designated by icl-0). We present a heatmap displaying the average differences in normalized AUC between learning paradigms in Figure \ref{fig:heatmap}, and a boxplot depicting the distribution of normalized AUC values for the different paradigms in Figure \ref{fig:auc-learning-paradigm}.

\subsection{Analysis of Intrinsic Dimension Curves}

\subsubsection{Differing Shapes of Intrinsic Dimension Curves}
We observe that the exact shape of the Intrinsic Dimension curves is highly dependent on the dataset. For some datasets, such as AG News, we observe a consistent "hunchback" shape, where the ID initially increases and then is progressively lower in the later layers of the model across all models and learning paradigms (Figure \ref{fig:comparison-ag_news}). This shape has been reported by previous work \citep{yin2024characterizing} in QA datasets. However, this pattern does not consistently hold across all models, datasets, and learning paradigms. For example, on the QQP dataset, we do not observe a consistent hunchback shape for icl-0, icl-1, or fine-tuning learning paradigms (Figure \ref{fig:comparison-qqp}). In contrast, prior work has shown that Convolutional Neural Networks \citep{ansuini2019intrinsic}, as well as Image Generation Transformers such as ImageGPT and Protein Language Models \citep{valeriani2024geometry}, exhibit consistent Intrinsic Dimension patterns across their layers for inputs of their respective data modalities. This difference suggests that LLMs encode data into more diverse manifolds in their representation space, potentially reflecting their generality and the complexity of their learning tasks compared to other neural networks.

\begin{figure}[t]  % [t] for top placement, can use [h] for here or [b] for bottom
    \centering
    \includegraphics[width=\linewidth]{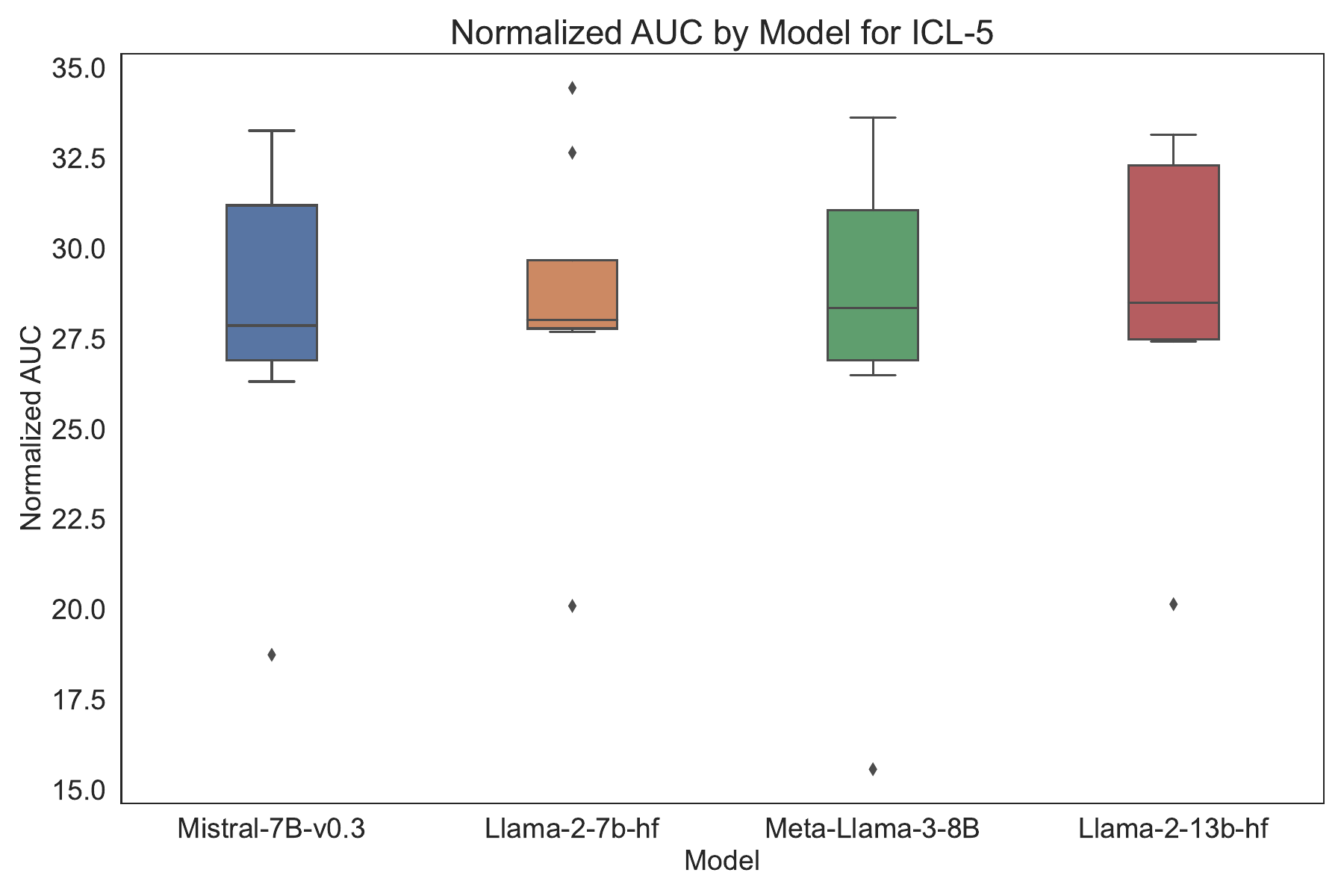}
    \caption{Boxplot displaying the distribution of normalized AUC values across datasets for each model in the ICL-5 shot setting. Each point corresponds to a (model, dataset) pair. The ID values lie in a narrow range, highlighting similarity in representation spaces across models.}
    \label{fig:auc_boxplot_icl}
\end{figure}

We also find that, within a specific learning paradigm, the range of normalized AUC values across datasets is similar for the four different models we tested, despite the fact that these models come from different families and have different embedding dimensions (e.g., Llama-2-13b has a hidden dimension of 5120, while the other three models have hidden dimensions of 4096). Figure \ref{fig:auc_boxplot_icl} depicts the range of normalized AUC values for the ICL-5 learning paradigm and shows that all values fall within a range of 20. We view this as evidence that different models may be generating representations with similar geometric complexity for a specific dataset, despite differences in model size or pre-training schemes. Similar boxplots for normalized AUC values from other experiments are included in Appendix \ref{sec: auc-model-finetune}. These findings are in agreement with results from \citet{cheng2023bridging}, which show that LLMs of different sizes and families create representations with similar ID values for a variety of text corpora.

\subsection{Comparing Performance of Different Learning Paradigms}

We found that models fine-tuned with 1k samples obtained the highest accuracy, while models performing ICL with 10 samples followed closely. This observation suggests that intrinsic dimension (ID) may not be directly related to accuracy: although fine-tuning with 1k samples yields ID values that remain closer to the baseline model, ICL models exhibit higher IDs yet achieve substantially lower accuracies. See Table \ref{tab:acc-results} for the average performance of each learning paradigm across the models and datasets tested.

\section{Summary}
We present a detailed analysis of the intrinsic dimension (ID) induced by the SFT and ICL learning paradigms. Our experiments reveal that the normalized AUC of ID curves sometimes decreases during the initial stages of SFT but generally increases during the later stages. 

Additionally, we observe that the normalized AUC of ID curves in ICL initially increases for small values of \( k \) (the number of demonstrations) but plateaus or slightly decreases as \( k \) increases further. Notably, the \( k \) value corresponding to the highest normalized AUC also achieves peak (or near-peak) accuracy, suggesting that ID may serve as a useful indicator for selecting the optimal number of demonstrations during ICL. 

Finally, our direct comparison of ID curves from ICL and SFT reveals that representations generated during ICL consistently yield higher ID curves compared to those from SFT on 1k samples, even though SFT with 1k samples achieves the highest overall performance. This analysis provides evidence that the two learning paradigms induce distinct representational structures in the embedding space, with ICL representations occupying higher-dimensional manifolds.

\section{Limitations}
In this study, we limit our analysis to models with sizes between 7B and 13B parameters. Future work may extend this investigation to models of different sizes. We also focus on datasets defined by narrowly focused tasks and do not consider datasets with long-form answers. Due to computational constraints, we perform fine-tuning only using LoRA adapters and do not explore the impacts of full fine-tuning on intrinsic dimension.

\bibliography{anthology,custom}
\bibliographystyle{acl_natbib}

\appendix

% \section{Accuracy Table Comparing Performance of Learning Paradigms}
% \label{sec:acc-table-box}
% \begin{table*}[ht]
% \centering
% \begin{tabular}{lcccccc}
% \hline
% \textbf{Dataset} & \textbf{ICL-0} & \textbf{ICL-1} & \textbf{ICL-2} & \textbf{ICL-5} & \textbf{ICL-10} & \textbf{Finetune 1K}\\
% \hline
% \textbf{SST-2} & 0.685 & 0.633 & 0.731 & 0.807 & 0.832 & \textbf{0.944} \\
% \textbf{CoLA} & 0.720 & 0.723 & 0.735 & 0.746 & 0.742 & \textbf{0.750} \\
% \textbf{QNLI} & 0.517 & 0.513 & 0.555 & 0.590 & 0.585 & \textbf{0.761} \\
% \textbf{QQP} & 0.417 & 0.462 & 0.485 & 0.508 & 0.519 & \textbf{0.707} \\
% \textbf{MNLI} & 0.374 & 0.367 & 0.387 & 0.414 & 0.431 & \textbf{0.676} \\
% \textbf{AG News} & 0.638 & 0.573 & 0.712 & 0.772 & 0.809 & \textbf{0.881} \\
% \textbf{Commonsense QA} & 0.199 & 0.375 & 0.417 & 0.470 & 0.492 & \textbf{0.500} \\
% \textbf{MMLU} & 0.449 & 0.488 & 0.511 & 0.524 & 0.531 & \textbf{0.542} \\
% \hline
% \end{tabular}
% \caption{Accuracy results for different datasets across in-context learning (ICL) settings and fine-tuning with 1K samples. Bold values represent the highest accuracy for each dataset.}
% \label{tab:acc-results}
% \end{table*}

\twocolumn[\section{In-Context Learning Experiments}\label{sec:detailed-icl}]

\FloatBarrier

\subsection{Llama-3-8B In-Context Learning Experiments}

\vspace{0.6cm} % Adjust spacing as needed
\noindent\begin{minipage}{\textwidth}
    \centering
    \includegraphics[width=\textwidth]{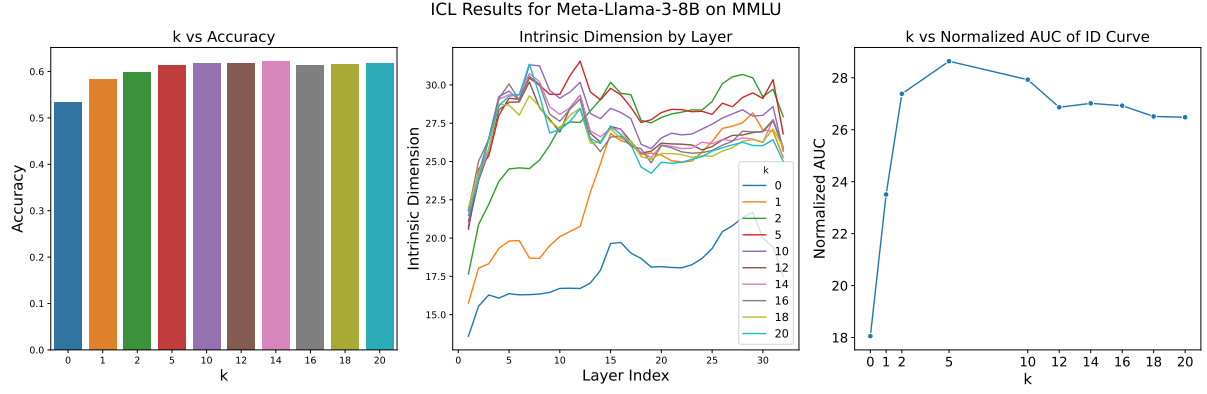}
    \captionof{figure}{ICL Experiment Results for Meta-Llama-3-8B on MMLU}
    \label{fig:icl-llama-8b-mmlu}
\end{minipage}

\FloatBarrier

\vspace{0.6cm}
\noindent\begin{minipage}{\textwidth}
    \centering
    \includegraphics[width=\textwidth]{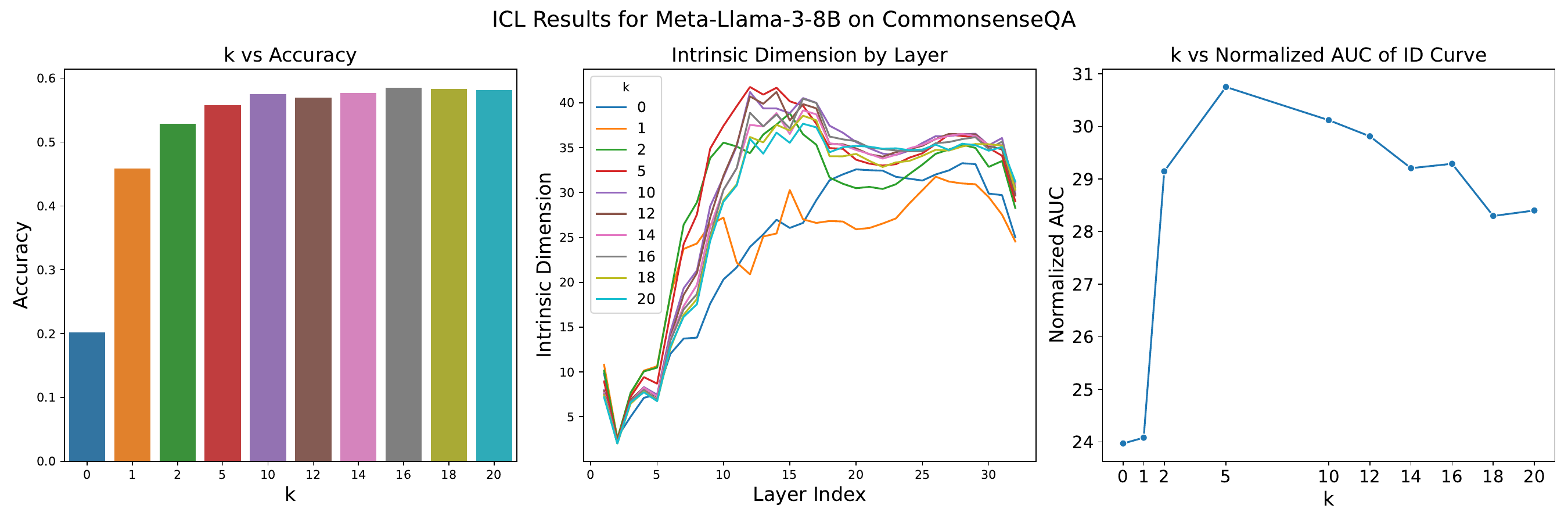}
    \captionof{figure}{ICL Experiment Results for Meta-Llama-3-8B on CommonsenseQA}
    \label{fig:icl-llama-8b-commonsense_qa}
\end{minipage}

\FloatBarrier

\vspace{0.6cm}
\noindent\begin{minipage}{\textwidth}
    \centering
    \includegraphics[width=\textwidth]{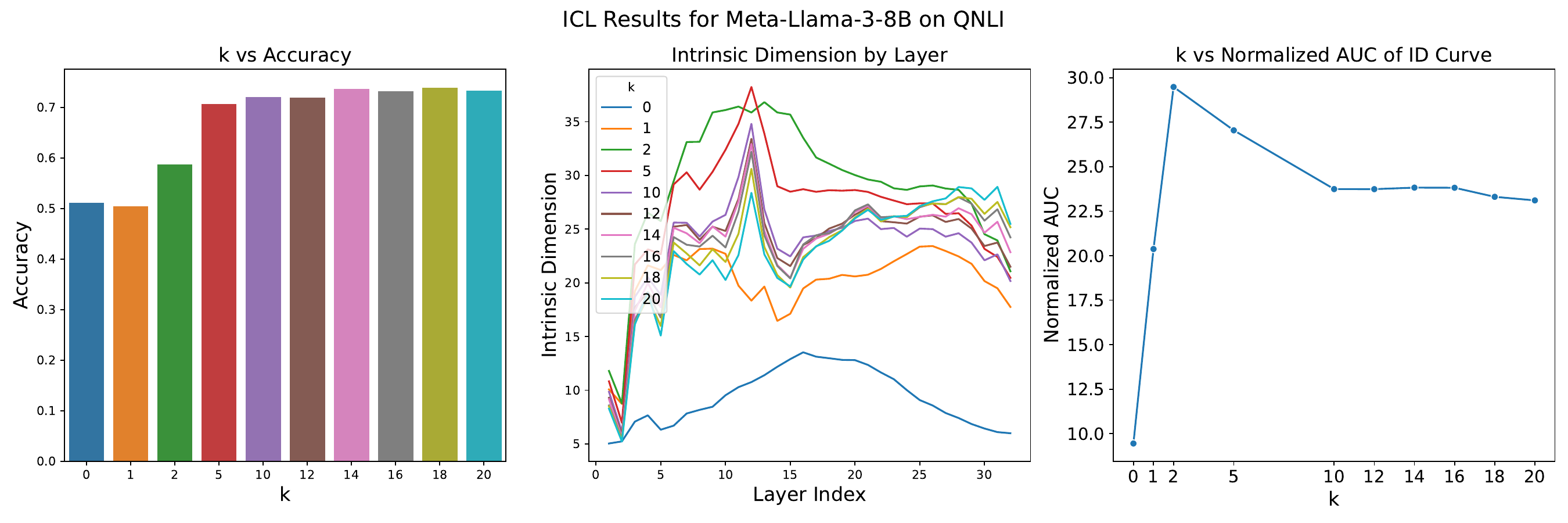}
    \captionof{figure}{ICL Experiment Results for Meta-Llama-3-8B on QNLI}
    \label{fig:icl-llama-8b-qnli}
\end{minipage}

\clearpage

\subsection{Llama-2-13b In-Context Learning Experiments}
\FloatBarrier

\vspace{0.6cm} % Adjust spacing as needed
\noindent\begin{minipage}{\textwidth}
    \centering
    \includegraphics[width=\textwidth]{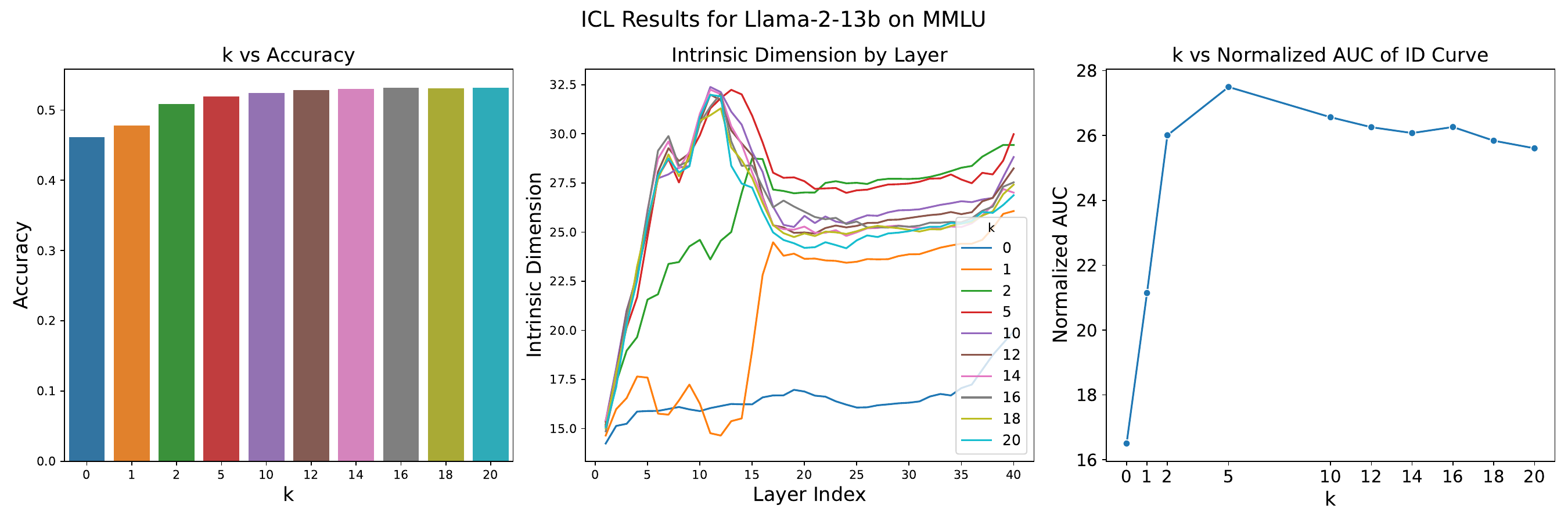}
    \captionof{figure}{ICL Experiment Results for Llama-2-13b on MMLU}
    \label{fig:icl-llama-8b-mmlu}
\end{minipage}

\FloatBarrier

\vspace{0.6cm}
\noindent\begin{minipage}{\textwidth}
    \centering
    \includegraphics[width=\textwidth]{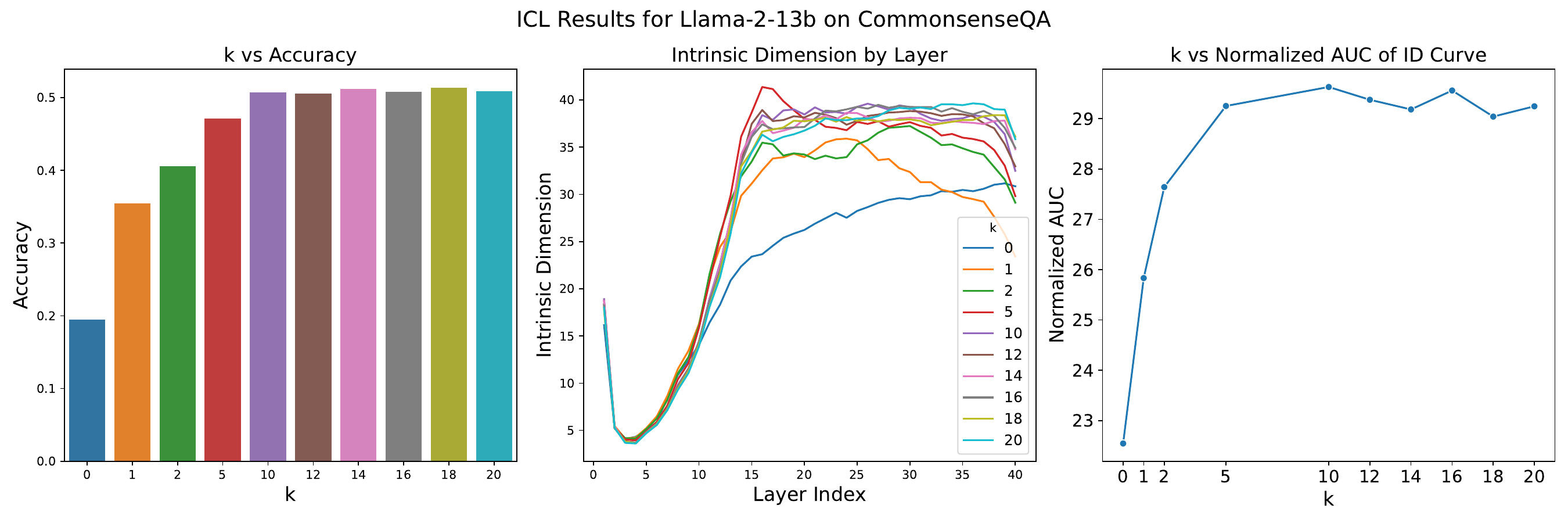}
    \captionof{figure}{ICL Experiment Results for Llama-2-13b on CommonsenseQA}
    \label{fig:icl-llama-8b-commonsense_qa}
\end{minipage}

\FloatBarrier

\vspace{0.6cm}
\noindent\begin{minipage}{\textwidth}
    \centering
    \includegraphics[width=\textwidth]{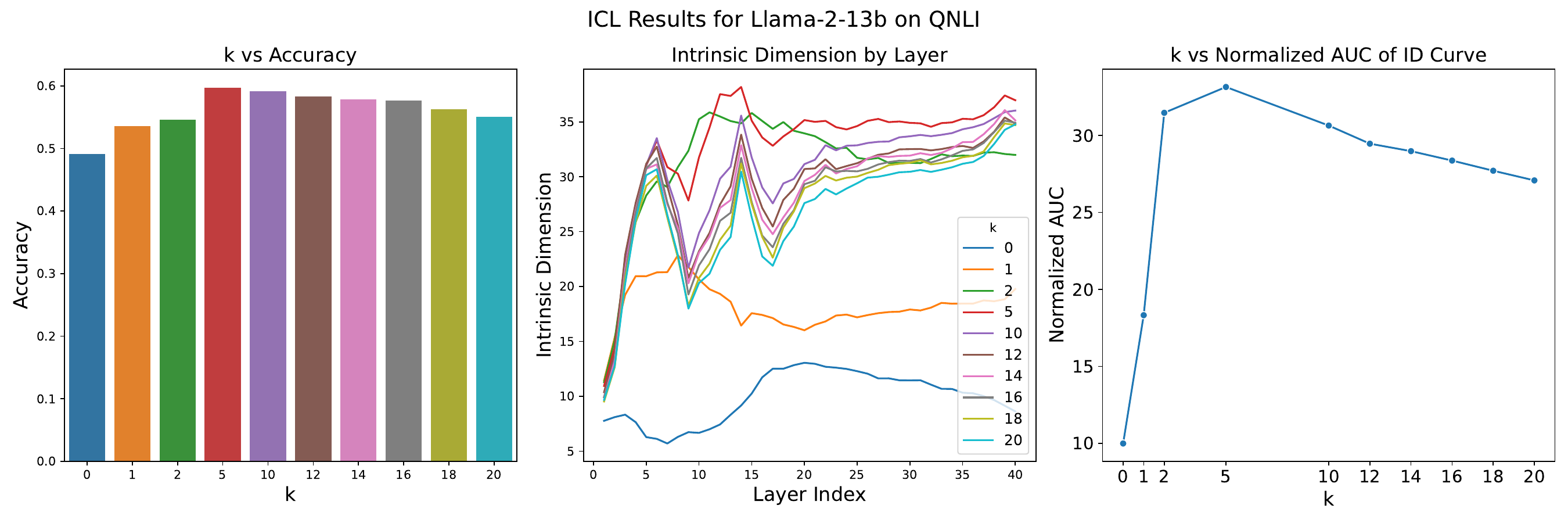}
    \captionof{figure}{ICL Experiment Results for Llama-2-13b on QNLI}
    \label{fig:icl-llama-8b-qnli}
\end{minipage}

\clearpage

\section{Supervised Fine-Tuning Experiments}
\label{sec:detailed-sft}

\subsection{Supervised Fine-Tuning Results for Llama-3-8B}

\FloatBarrier

\vspace{0.6cm} % Adjust spacing as needed
\noindent\begin{minipage}{\textwidth}
    \centering
    \includegraphics[width=\textwidth]{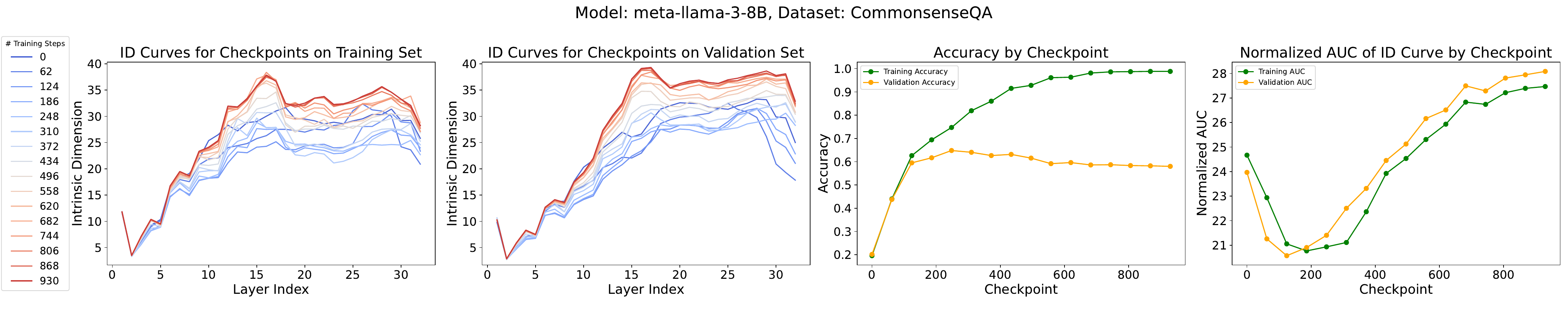}
    \captionof{figure}{Supervised Fine-Tuning Results for Llama-3-8B on Commonsense QA}
    \label{fig:icl-llama-8b-mmlu}
\end{minipage}

\vspace{0.6cm} % Adjust spacing as needed
\noindent\begin{minipage}{\textwidth}
    \centering
    \includegraphics[width=\textwidth]{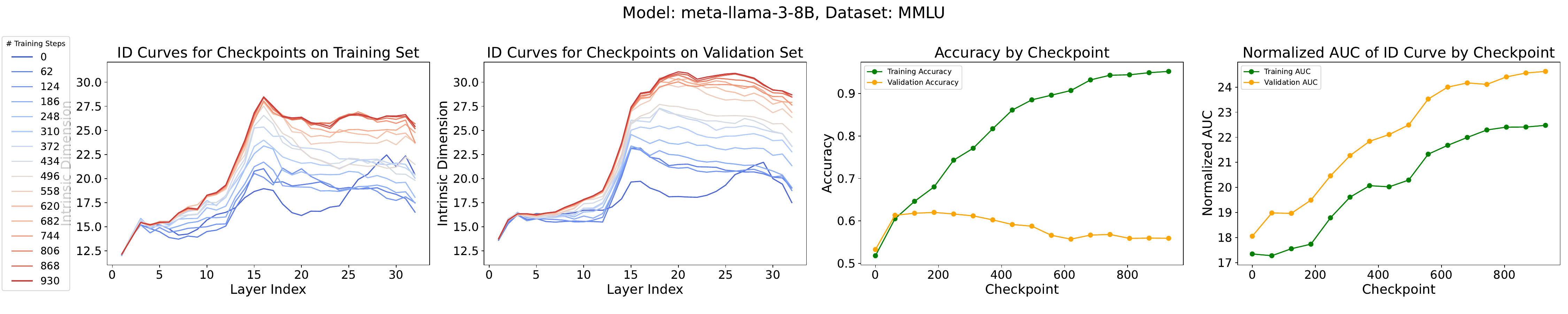}
    \captionof{figure}{Supervised Fine-Tuning Results for Llama-3-8B on MMLU}
    \label{fig:sft-llama-8b-mmlu}
\end{minipage}

\vspace{0.6cm} % Adjust spacing as needed
\noindent\begin{minipage}{\textwidth}
    \centering
    \includegraphics[width=\textwidth]{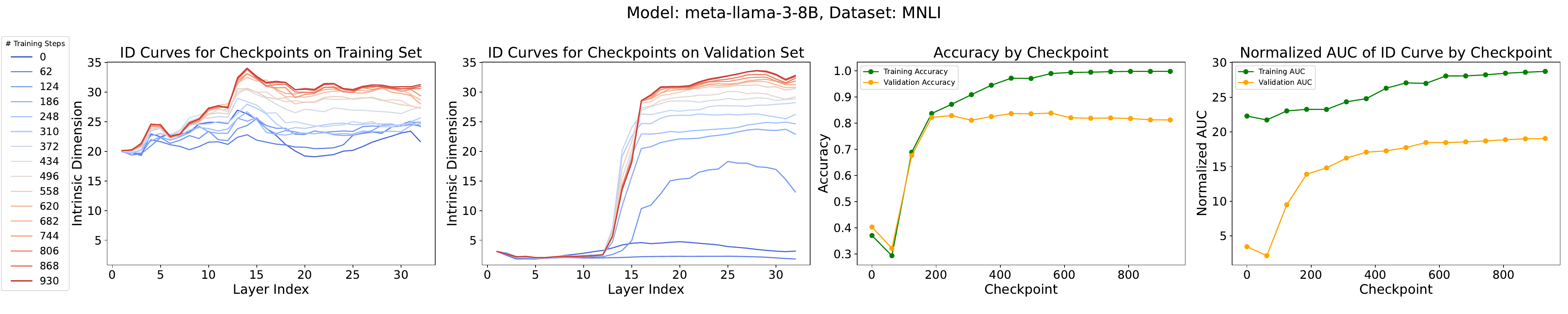}
    \captionof{figure}{Supervised Fine-Tuning Results for Llama-3-8B on MNLI}
    \label{fig:sft-llama-8b-mnli}
\end{minipage}

\vspace{0.6cm} % Adjust spacing as needed
\noindent\begin{minipage}{\textwidth}
    \centering
    \includegraphics[width=\textwidth]{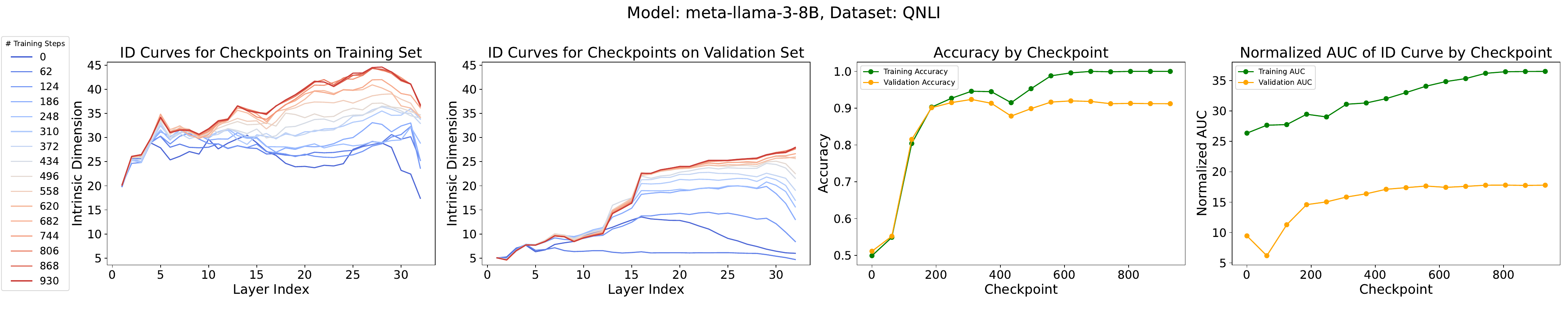}
    \captionof{figure}{Supervised Fine-Tuning Results for Llama-3-8B on QNLI}
    \label{fig:sft-llama-8b-qnli}
\end{minipage}

\vspace{0.6cm} % Adjust spacing as needed
\noindent\begin{minipage}{\textwidth}
    \centering
    \includegraphics[width=\textwidth]{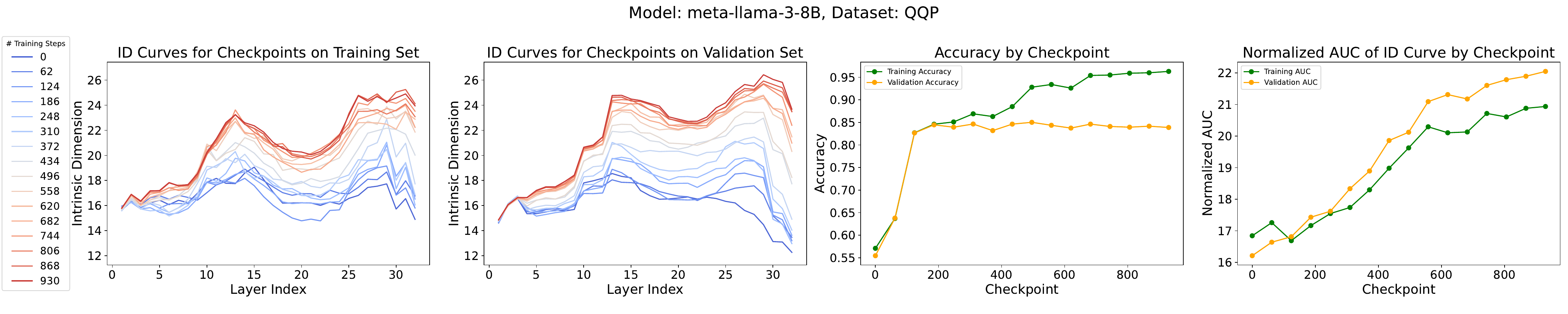}
    \captionof{figure}{Supervised Fine-Tuning Results for Llama-3-8B on QQP}
    \label{fig:sft-llama-8b-qqp}
\end{minipage}

\clearpage

\vspace{0.6cm} % Adjust spacing as needed
\noindent\begin{minipage}{\textwidth}
    \centering
    \includegraphics[width=\textwidth]{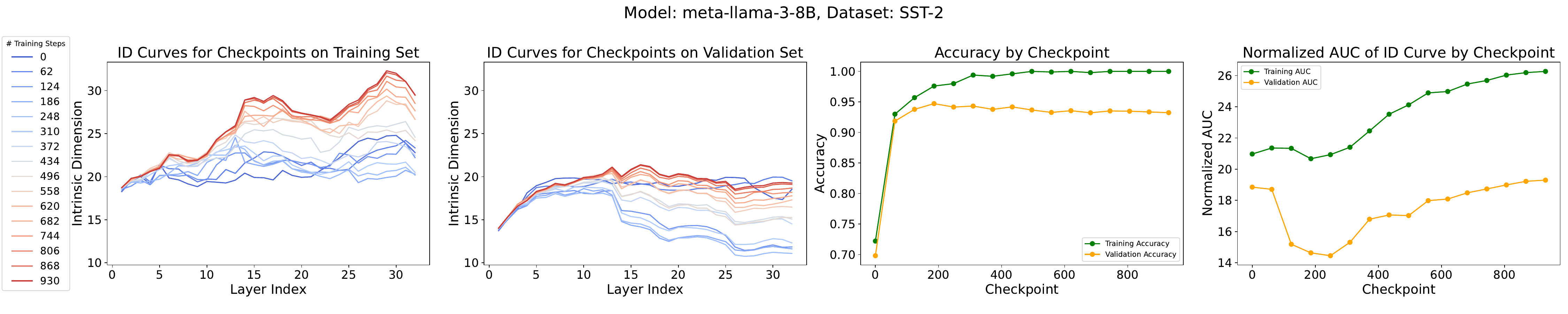}
    \captionof{figure}{Supervised Fine-Tuning Results for Llama-3-8B on SST-2}
    \label{fig:sft-llama-8b-sst2}
\end{minipage}

\vspace{0.6cm} % Adjust spacing as needed
\noindent\begin{minipage}{\textwidth}
    \centering
    \includegraphics[width=\textwidth]{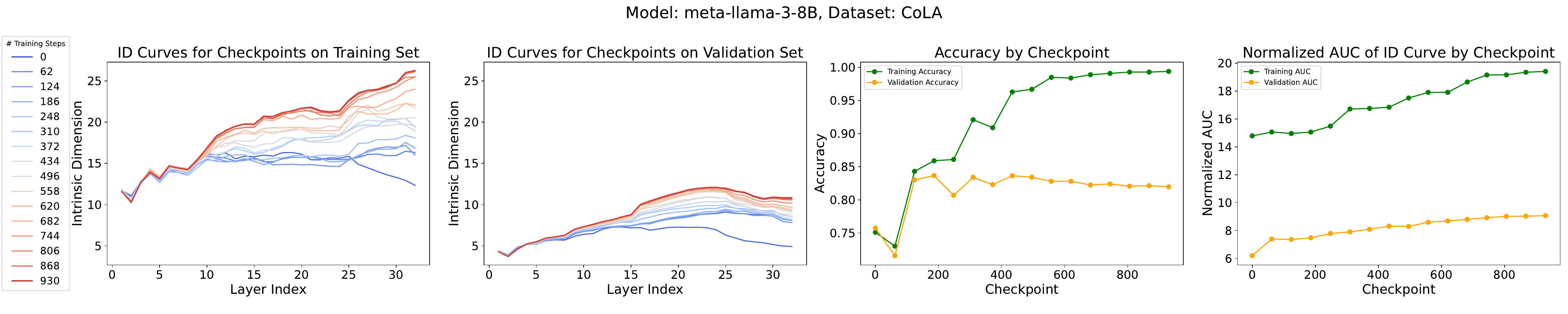}
    \captionof{figure}{Supervised Fine-Tuning Results for Llama-3-8B on CoLA}
    \label{fig:sft-llama-8b-cola}
\end{minipage}

\vspace{0.6cm} % Adjust spacing as needed
\noindent\begin{minipage}{\textwidth}
    \centering
    \includegraphics[width=\textwidth]{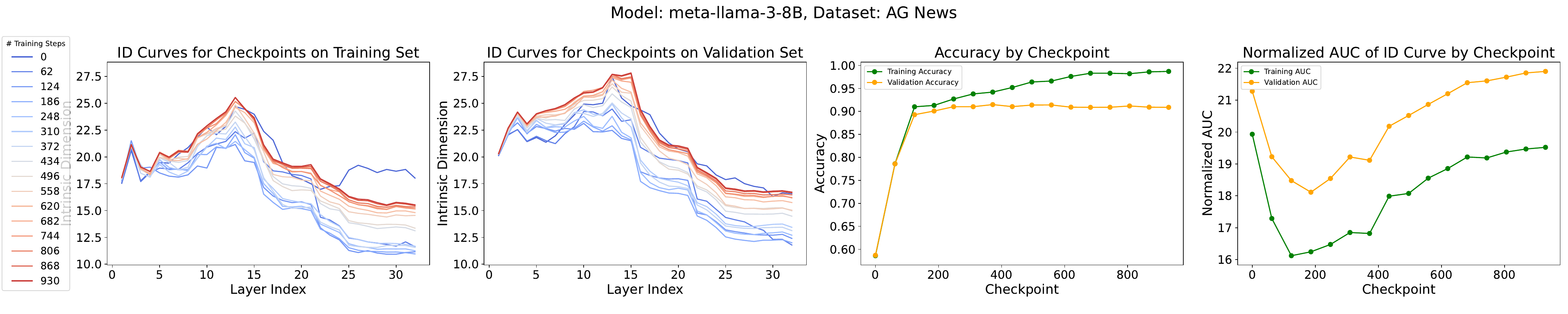}
    \captionof{figure}{Supervised Fine-Tuning Results for Llama-3-8B on AG News}
    \label{fig:sft-llama-8b-ag_news}
    \end{minipage}

\vspace{1.25cm} % Adjust spacing as needed

\subsection{Supervised Fine-Tuning Results for Llama-2-13B}

\vspace{0.6cm} % Adjust spacing as needed
\noindent\begin{minipage}{\textwidth}
    \centering
    \includegraphics[width=\textwidth]{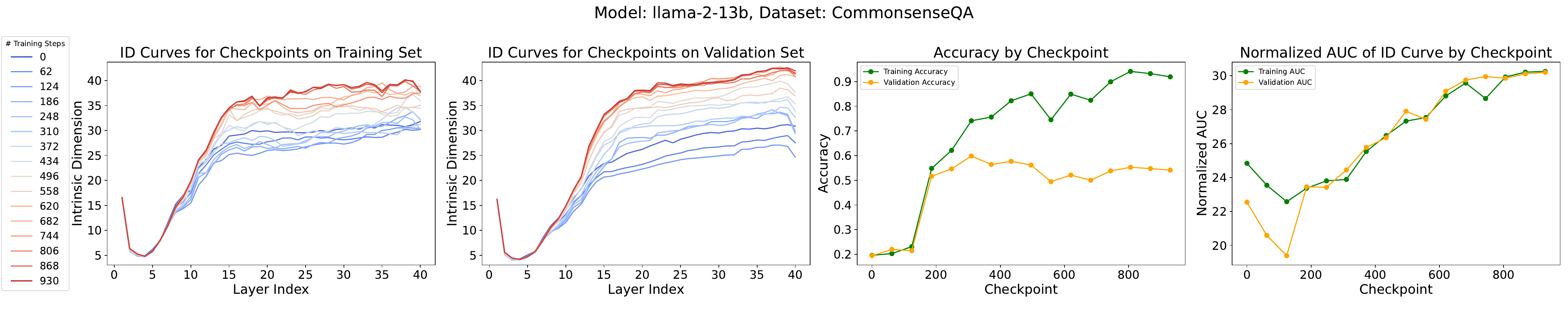}
    \captionof{figure}{Supervised Fine-Tuning Results for Llama-2-13B on Commonsense QA}
    \label{fig:sft-llama-2-13b-commonsense_qa}
\end{minipage}

\vspace{0.6cm} % Adjust spacing as needed
\noindent\begin{minipage}{\textwidth}
    \centering
    \includegraphics[width=\textwidth]{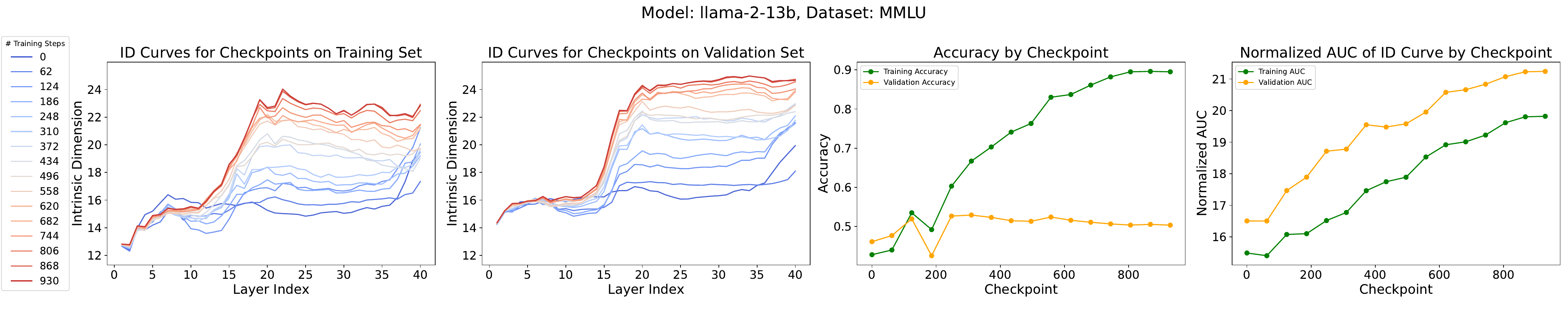}
    \captionof{figure}{Supervised Fine-Tuning Results for Llama-2-13B on MMLU}
    \label{fig:sft-llama-2-13b-mmlu}
\end{minipage}

\clearpage

\vspace{0.6cm} % Adjust spacing as needed
\noindent\begin{minipage}{\textwidth}
    \centering
    \includegraphics[width=\textwidth]{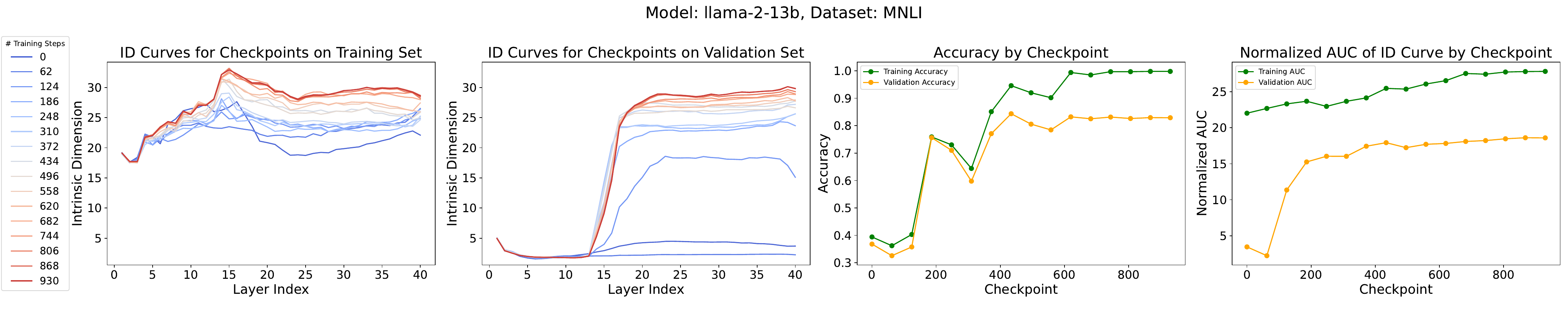}
    \captionof{figure}{Supervised Fine-Tuning Results for Llama-2-13B on MNLI}
    \label{fig:sft-llama-2-13b-mnli}
\end{minipage}

\vspace{0.6cm} % Adjust spacing as needed
\noindent\begin{minipage}{\textwidth}
    \centering
    \includegraphics[width=\textwidth]{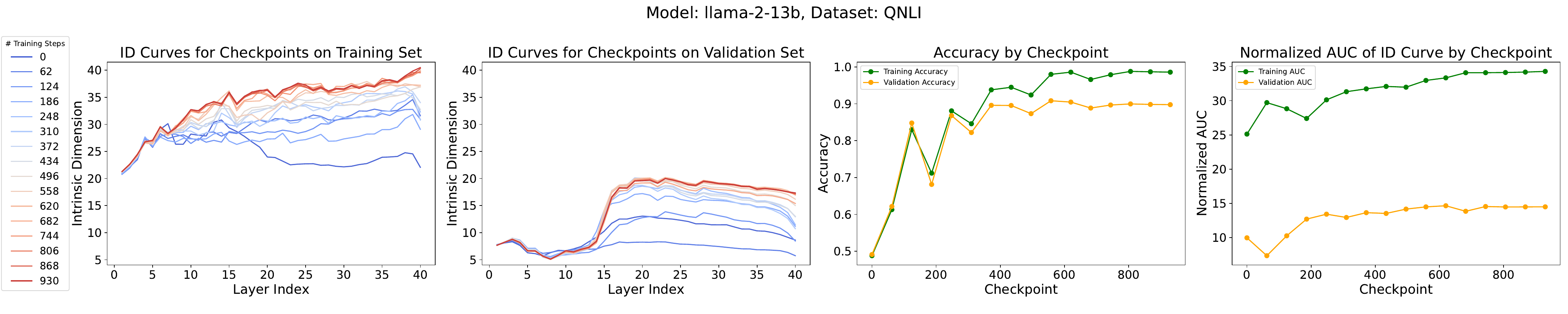}
    \captionof{figure}{Supervised Fine-Tuning Results for Llama-2-13B on QNLI}
    \label{fig:sft-llama-2-13b-qnli}
\end{minipage}

\vspace{0.6cm} % Adjust spacing as needed
\noindent\begin{minipage}{\textwidth}
    \centering
    \includegraphics[width=\textwidth]{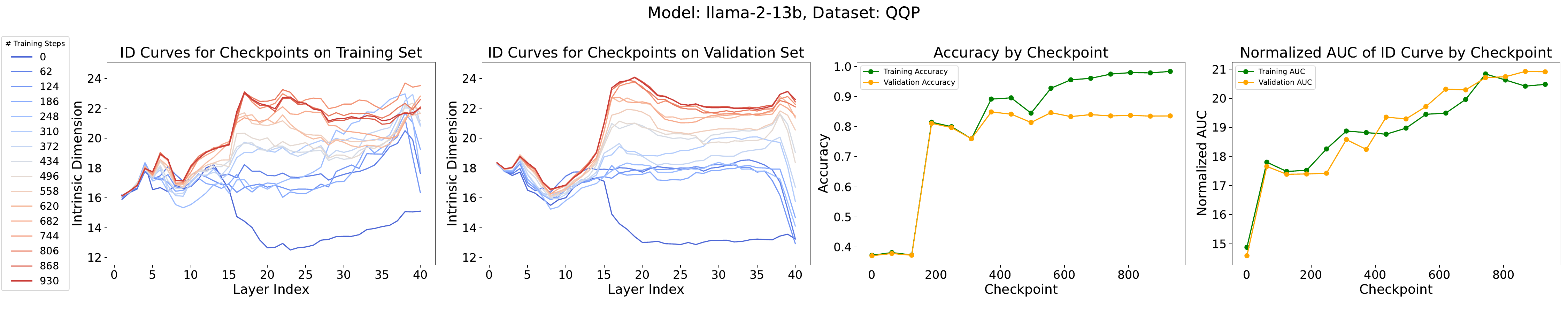}
    \captionof{figure}{Supervised Fine-Tuning Results for Llama-2-13B on QQP}
    \label{fig:sft-llama-2-13b-qqp}
\end{minipage}

\vspace{0.6cm} % Adjust spacing as needed
\noindent\begin{minipage}{\textwidth}
    \centering
    \includegraphics[width=\textwidth]{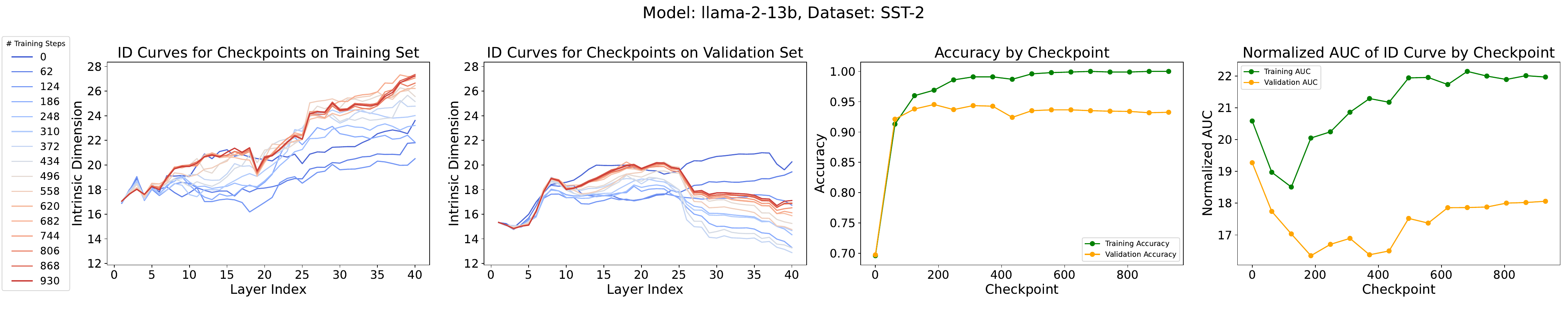}
    \captionof{figure}{Supervised Fine-Tuning Results for Llama-2-13B on SST-2}
    \label{fig:sft-llama-2-13b-sst2}
\end{minipage}

\vspace{0.6cm} % Adjust spacing as needed
\noindent\begin{minipage}{\textwidth}
    \centering
    \includegraphics[width=\textwidth]{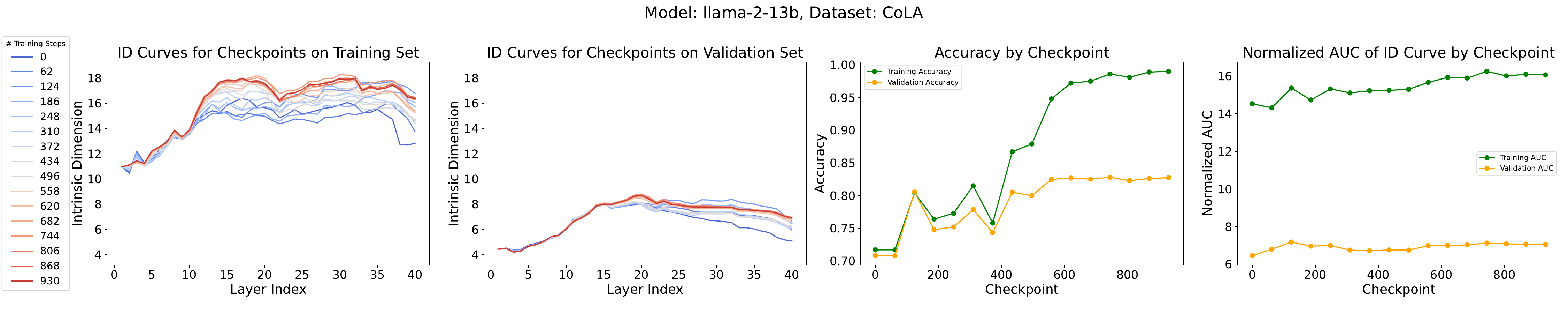}
    \captionof{figure}{Supervised Fine-Tuning Results for Llama-2-13B on CoLA}
    \label{fig:sft-llama-2-13b-cola}
\end{minipage}

\clearpage

\vspace{0.6cm} % Adjust spacing as needed
\noindent\begin{minipage}{\textwidth}
    \centering
    \includegraphics[width=\textwidth]{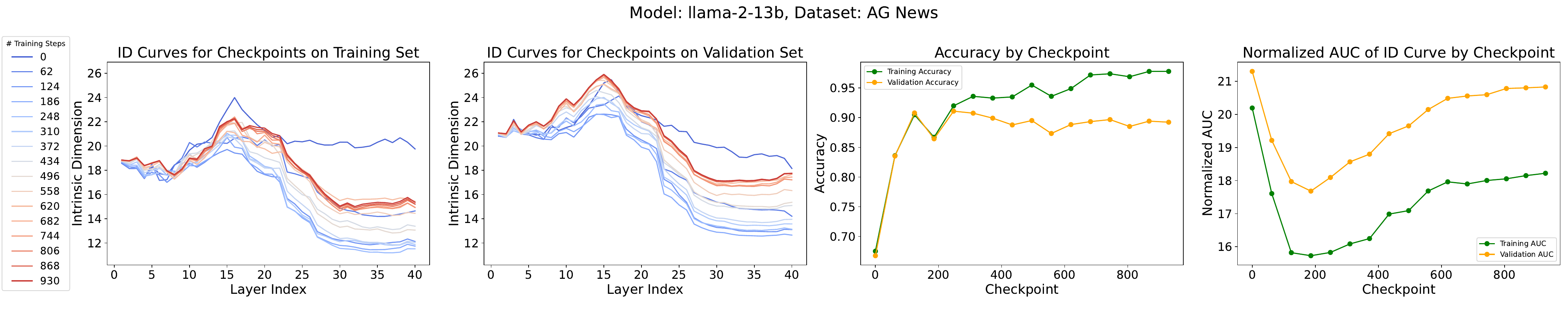}
    \captionof{figure}{Supervised Fine-Tuning Results for Llama-2-13B on AG News}
    \label{fig:sft-llama-2-13b-ag_news}
\end{minipage}

\clearpage

\twocolumn[\section{Comparisons of Supervised Fine-Tuning and In-Context Learning}]

\vspace{0.6cm} % Adjust spacing as needed
\noindent\begin{minipage}{\textwidth}
    \centering
    \includegraphics[width=\textwidth]{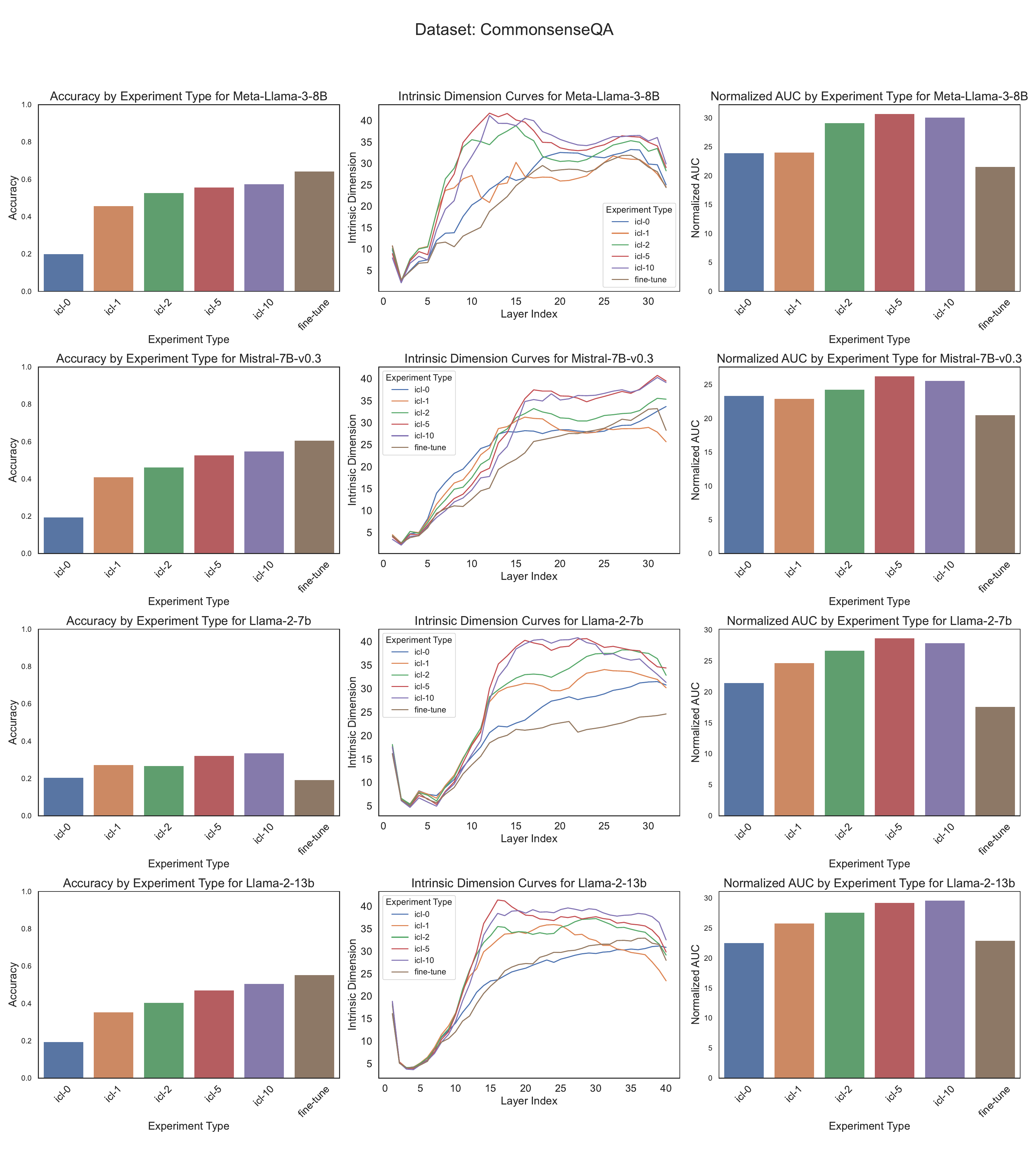}
    \captionof{figure}{Comparison of Experimental Results for Commonsense QA}
    \label{fig:comparison-commonsense_qa}
\end{minipage}

\begin{figure*}[t]
    \centering
    \includegraphics[width=\textwidth]{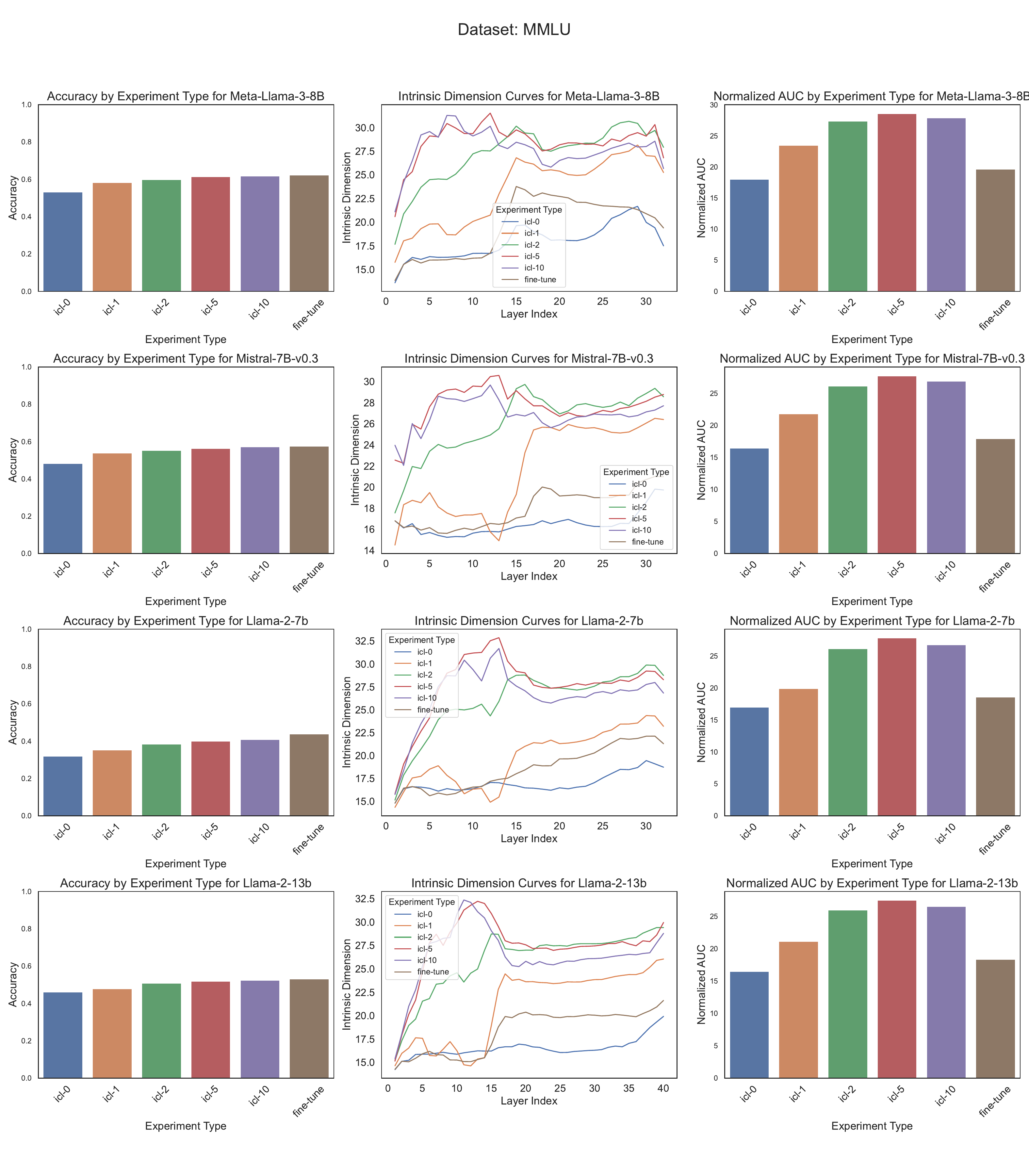}
    \caption{Comparison of Experimental Results for MMLU}
    \label{fig:comparison-mmlu}
\end{figure*}

\begin{figure*}[t]
    \centering
    \includegraphics[width=\textwidth]{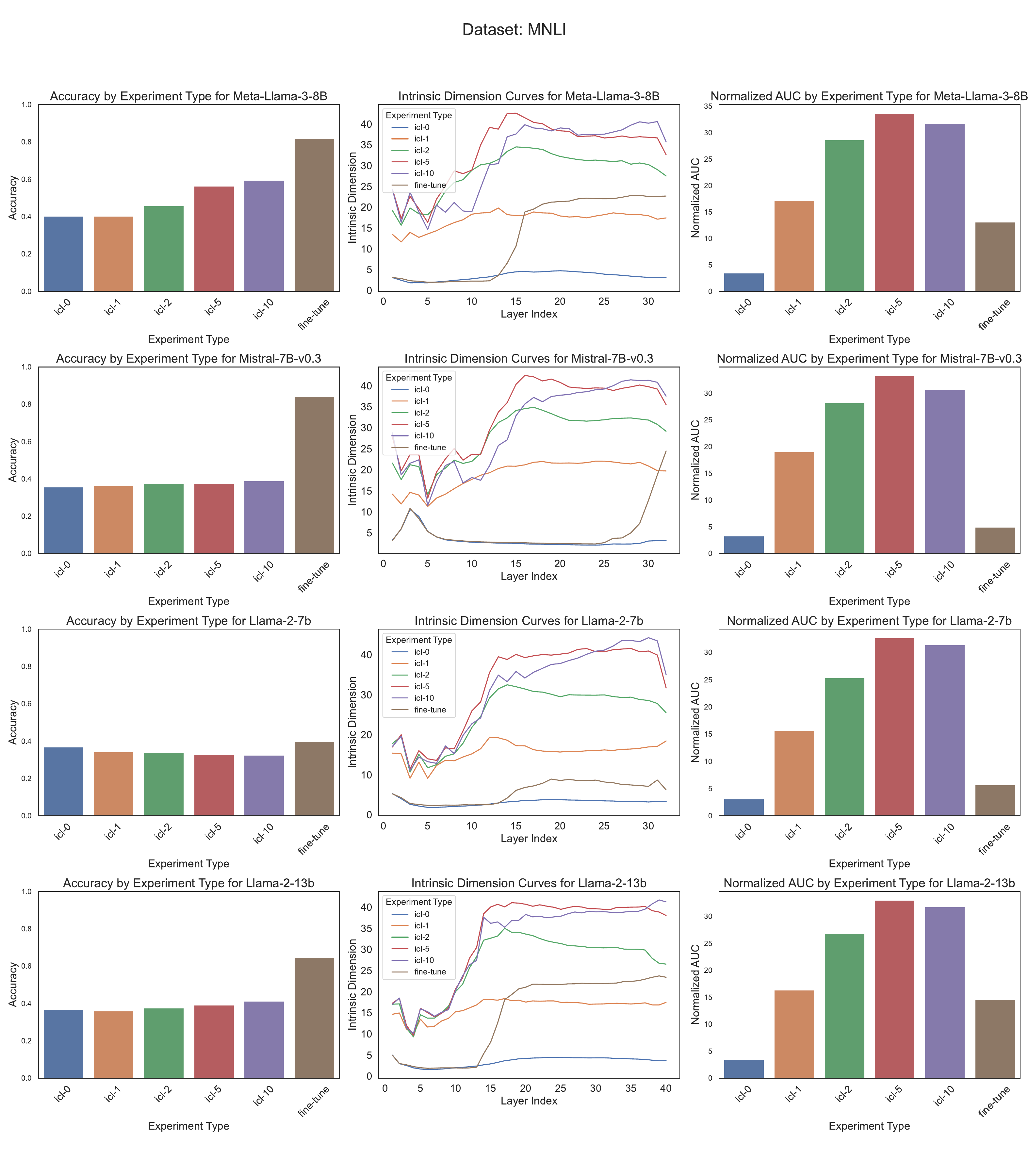}
    \caption{Comparison of Experimental Results for MNLI}
    \label{fig:comparison-mnli}
\end{figure*}

\begin{figure*}[t]
    \centering
    \includegraphics[width=\textwidth]{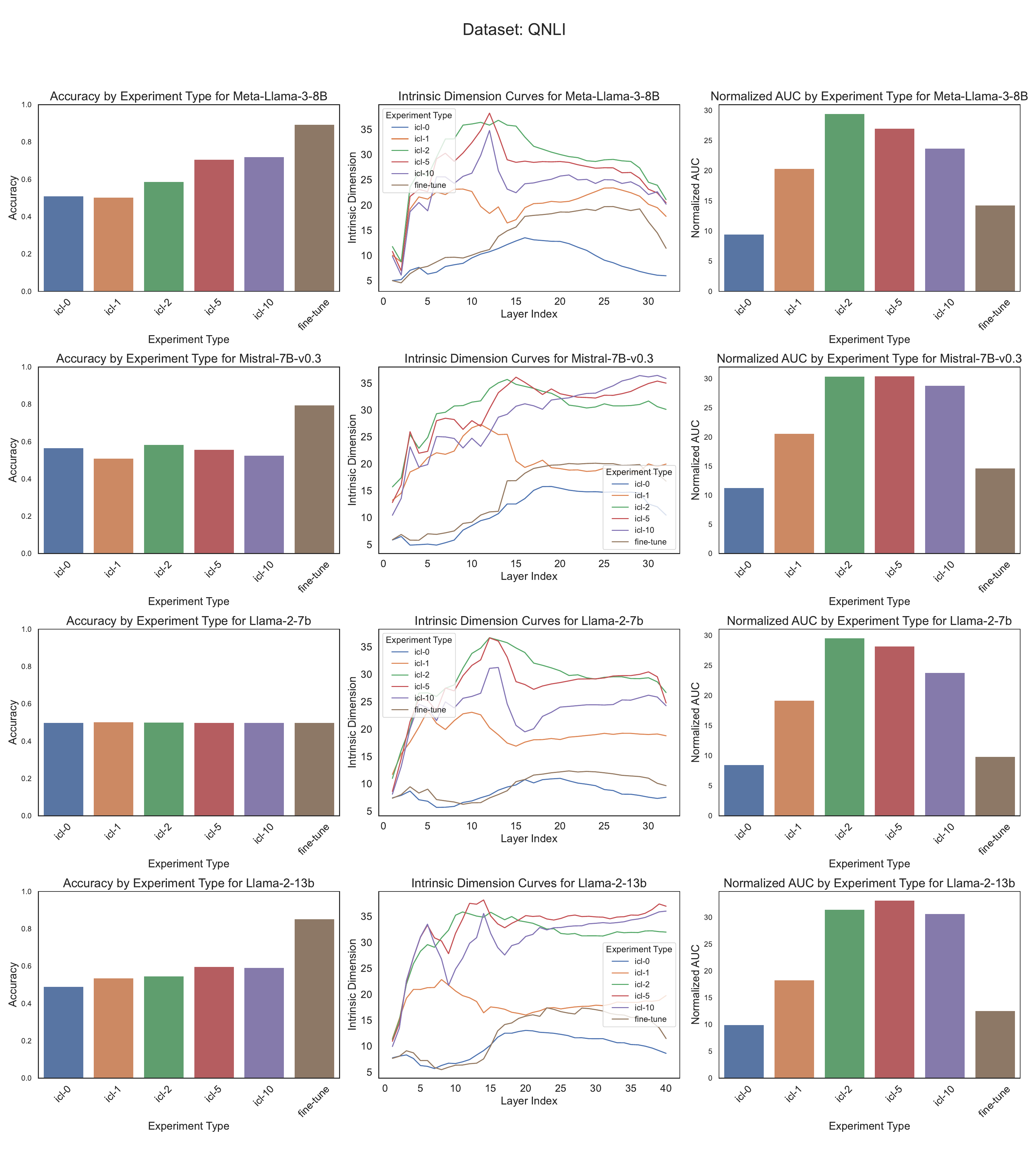}
    \caption{Comparison of Experimental Results for QNLI}
    \label{fig:comparison-qnli}
\end{figure*}

\begin{figure*}[t]
    \centering
    \includegraphics[width=\textwidth]{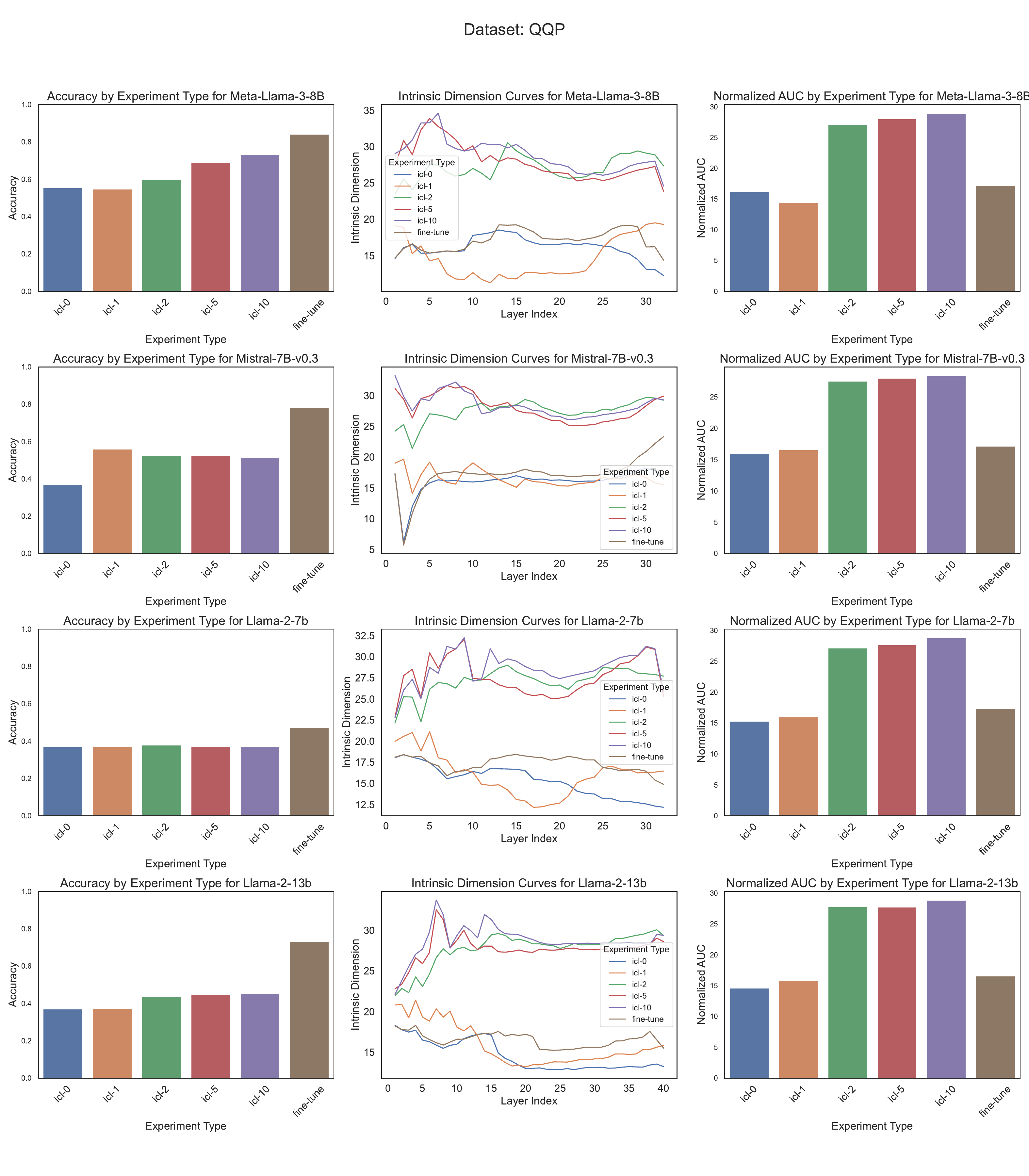}
    \caption{Comparison of Experimental Results for QQP}
    \label{fig:comparison-qqp}
\end{figure*}

\begin{figure*}[t]
    \centering
    \includegraphics[width=\textwidth]{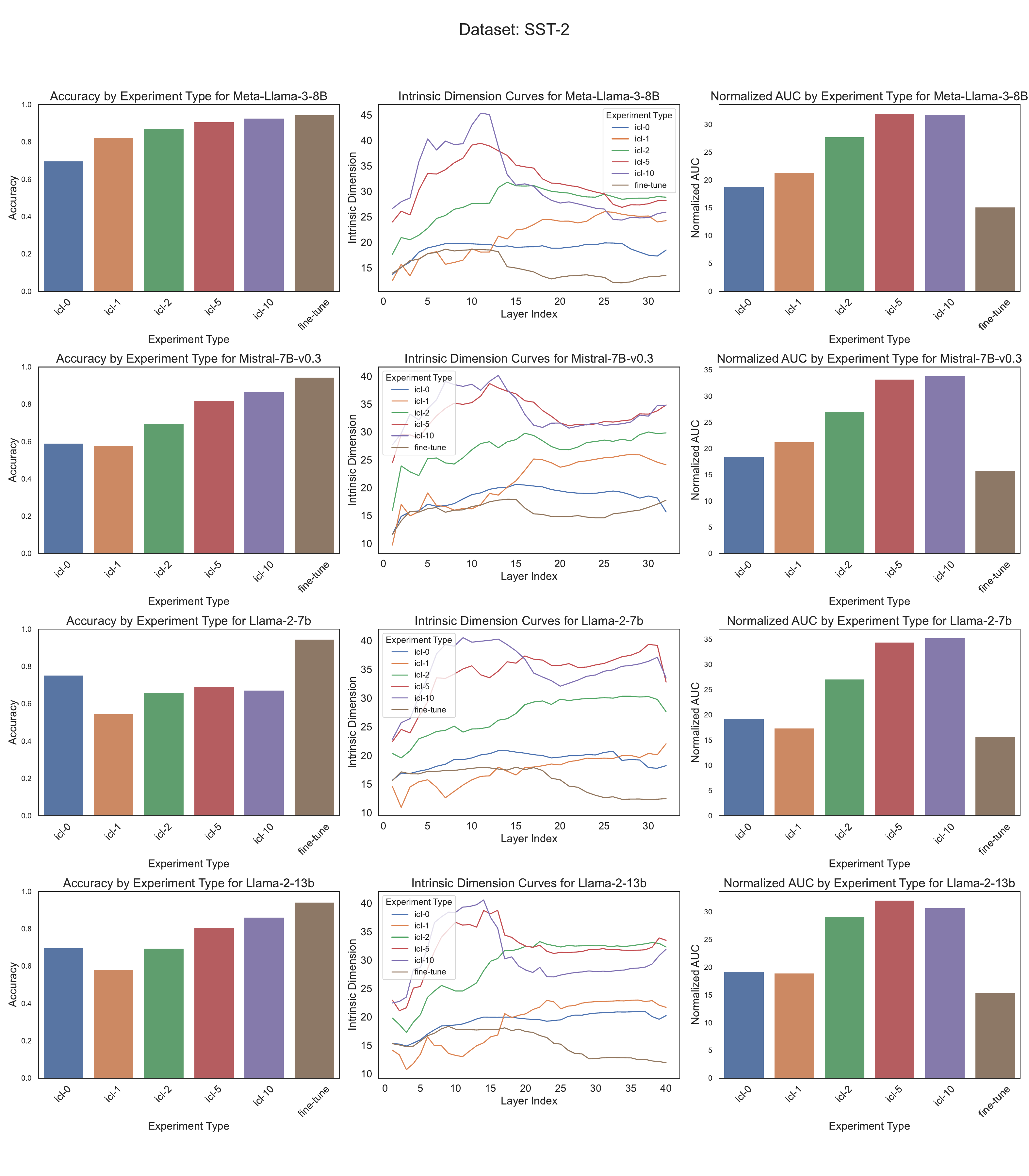}
    \caption{Comparison of Experimental Results for SST-2}
    \label{fig:comparison-sst2}
\end{figure*}

\begin{figure*}[t]
    \centering
    \includegraphics[width=\textwidth]{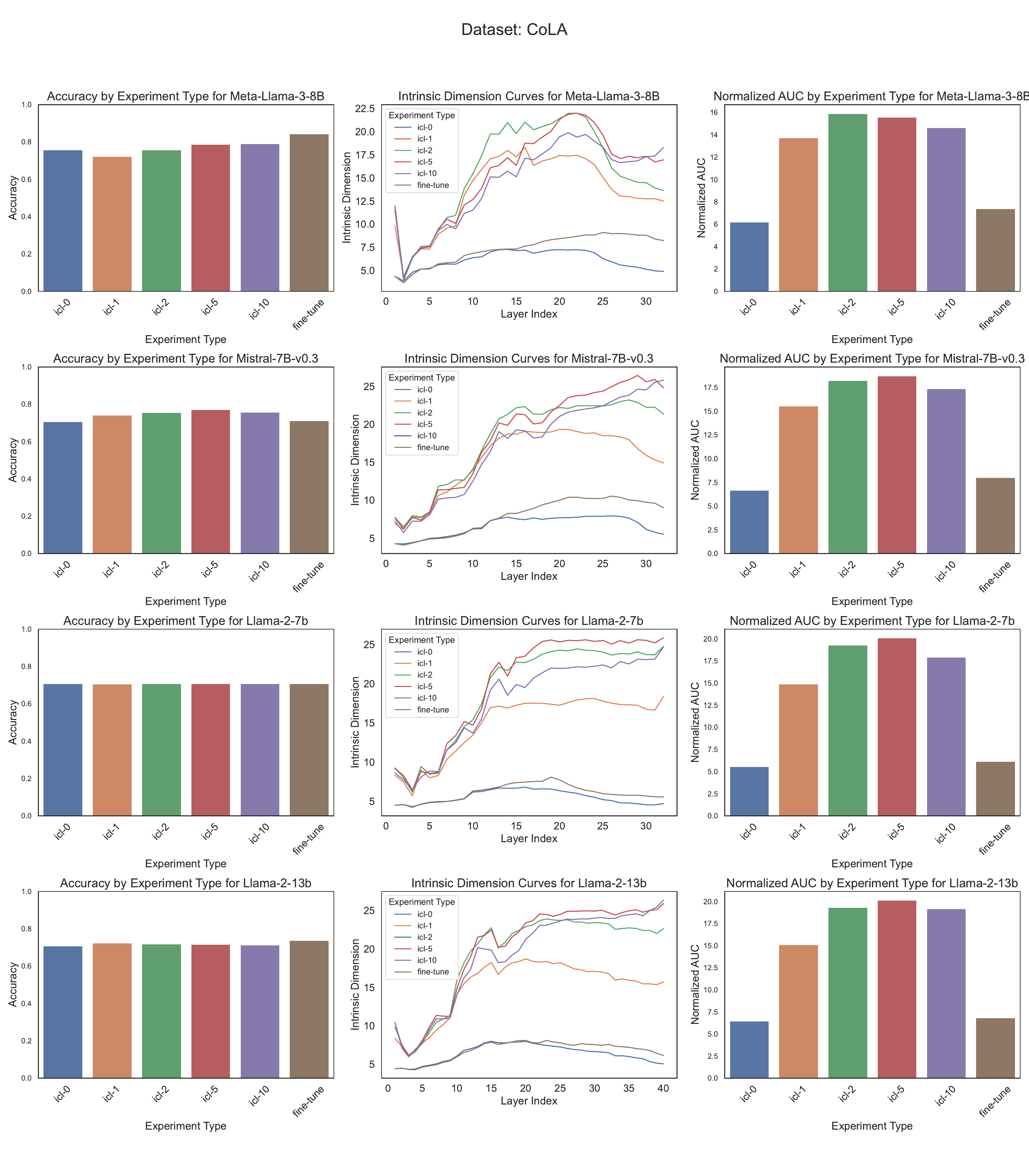}
    \caption{Comparison of Experimental Results for CoLA}
    \label{fig:comparison-cola}
\end{figure*}

\begin{figure*}[t]
    \centering
    \includegraphics[width=\textwidth]{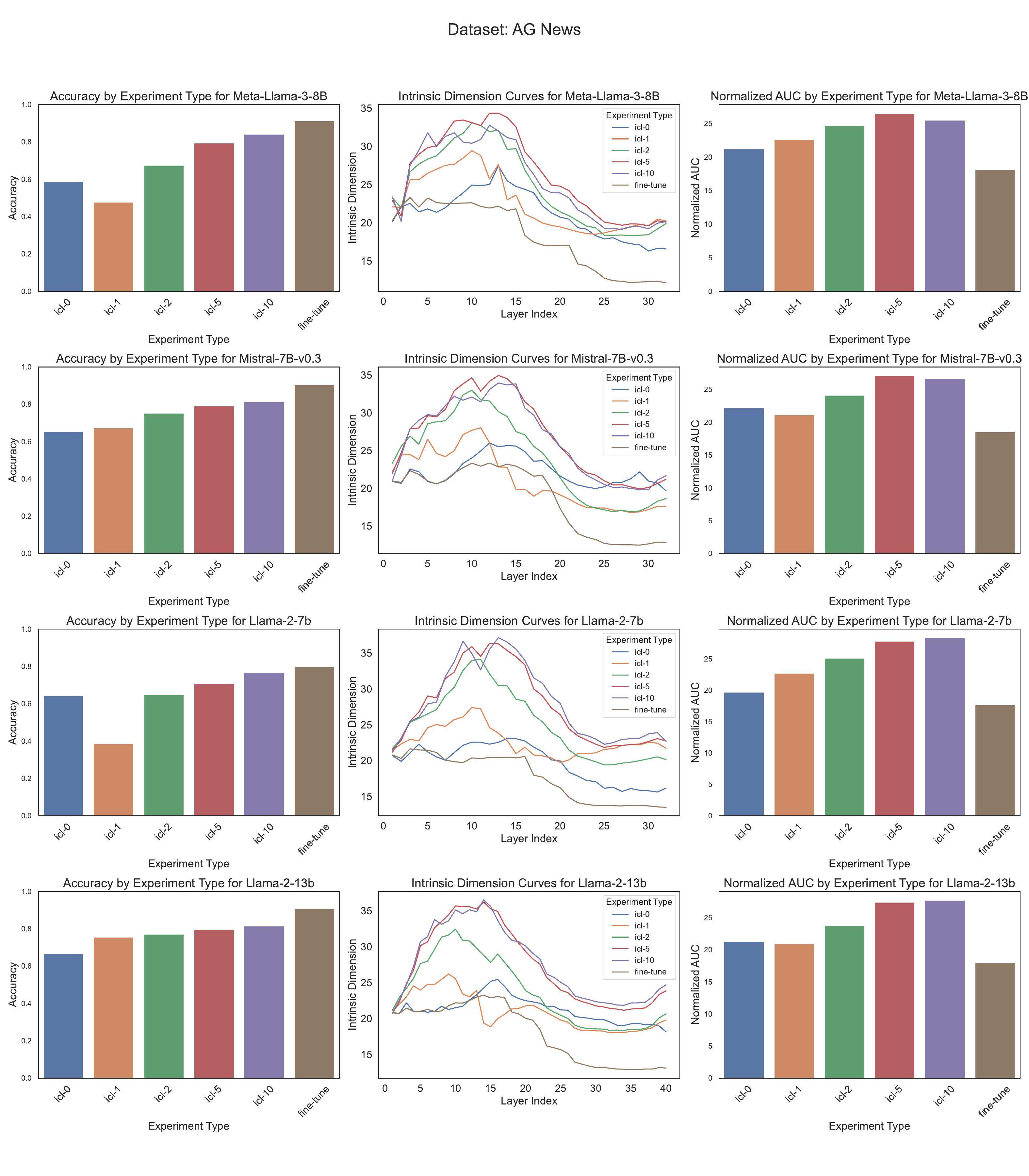}
    \caption{Comparison of Experimental Results for AG News}
    \label{fig:comparison-ag_news}
\end{figure*}
\clearpage

\twocolumn[\section{ICL Experiment Results with Unique Demonstrations}\label{sec: de-duplication}]

\vspace{0.8cm} % Adjust spacing as needed
\noindent\begin{minipage}{\textwidth}
    \centering
    \includegraphics[width=\textwidth]{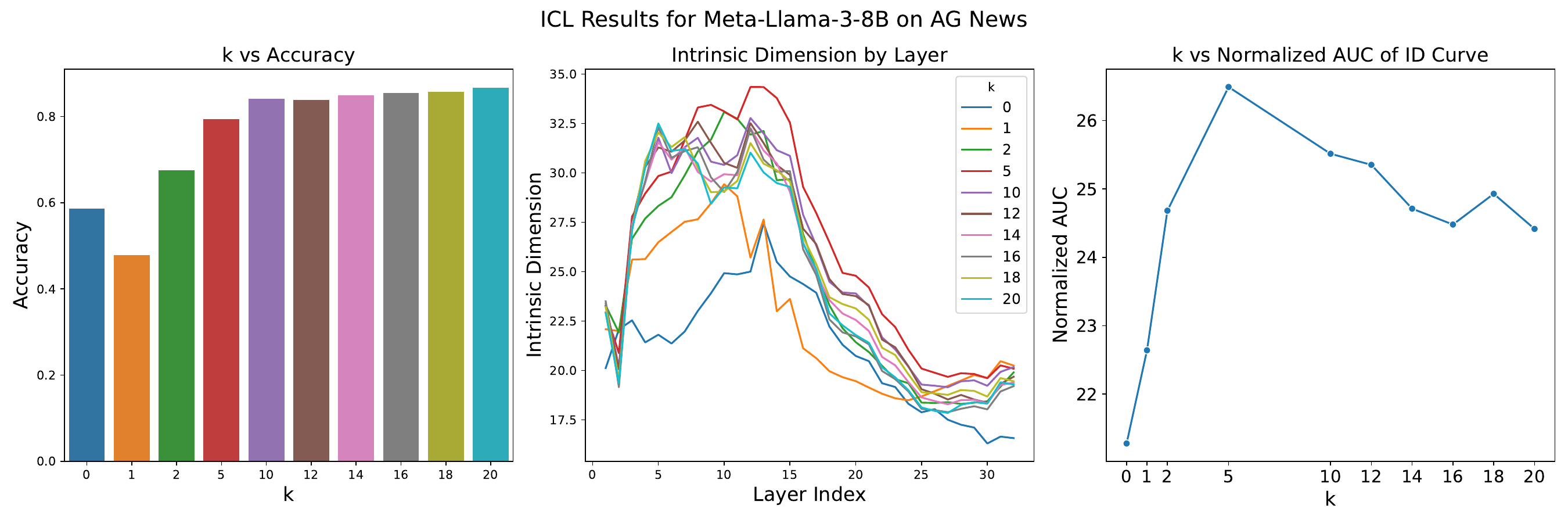}
    \captionof{figure}{ICL Experiment Results with Unique Demonstrations on AGNews Dataset}
    \label{fig:icl-dedup-ag_news}
\end{minipage}

\vspace{0.6cm} % Adjust spacing as needed
\noindent\begin{minipage}{\textwidth}
    \centering
    \includegraphics[width=\textwidth]{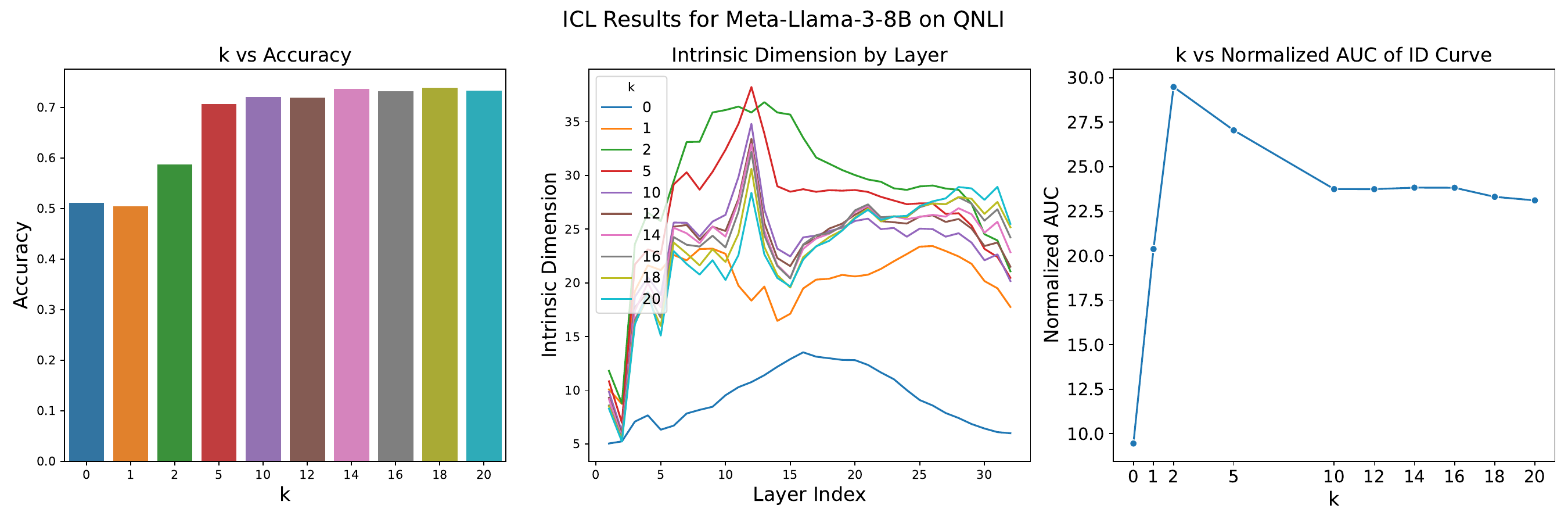}
    \captionof{figure}{ICL Experiment Results with Unique Demonstrations on QNLI Dataset}
    \label{fig:icl-dedup-qnli}
\end{minipage}

\vspace{0.6cm} % Adjust spacing as needed
\noindent\begin{minipage}{\textwidth}
    \centering
    \includegraphics[width=\textwidth]{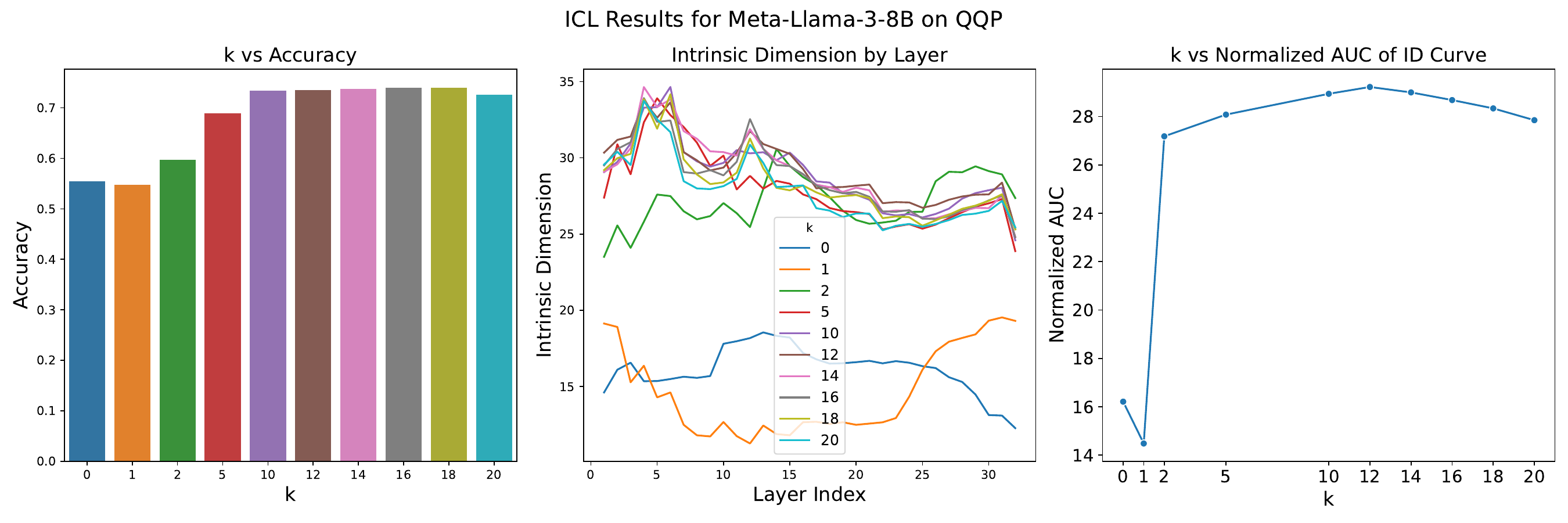}
    \captionof{figure}{ICL Experiment Results with Unique Demonstrations on QQP Dataset}
    \label{fig:icl-dedup-qqp}
\end{minipage}

% \label{sec: de-duplication}
% \begin{figure*}[t]
%     \centering
%     \includegraphics[width=\textwidth]{meta-llama-3-8B-ag_news-deduped.pdf}
%     \caption{ICL Experiment Results with Unique Demonstrations on AGNews Dataset}
%     \label{fig:icl-dedup-ag_news}
% \end{figure*}

% \begin{figure*}[h!]
%     \centering
%     \includegraphics[width=\textwidth]{meta-llama-3-8B-qnli-deduped.pdf}
%     \caption{ICL Experiment Results with Unique Demonstrations on QNLI Dataset}
%     \label{fig:icl-dedup-qnli}
% \end{figure*}

% \begin{figure*}[h!]
%     \centering
%     \includegraphics[width=\textwidth]{meta-llama-3-8B-qqp-deduped.pdf}
%     \caption{ICL Experiment Results with Unique Demonstrations on QQP Dataset}
%     \label{fig:icl-dedup-qqp}
% \end{figure*}

\clearpage

\twocolumn[\section{Normalized AUC Boxplot by Model for All Learning Paradigms}]
\label{sec: auc-model-finetune}

\begin{figure}[H]
    \centering
    \includegraphics[width=\linewidth]{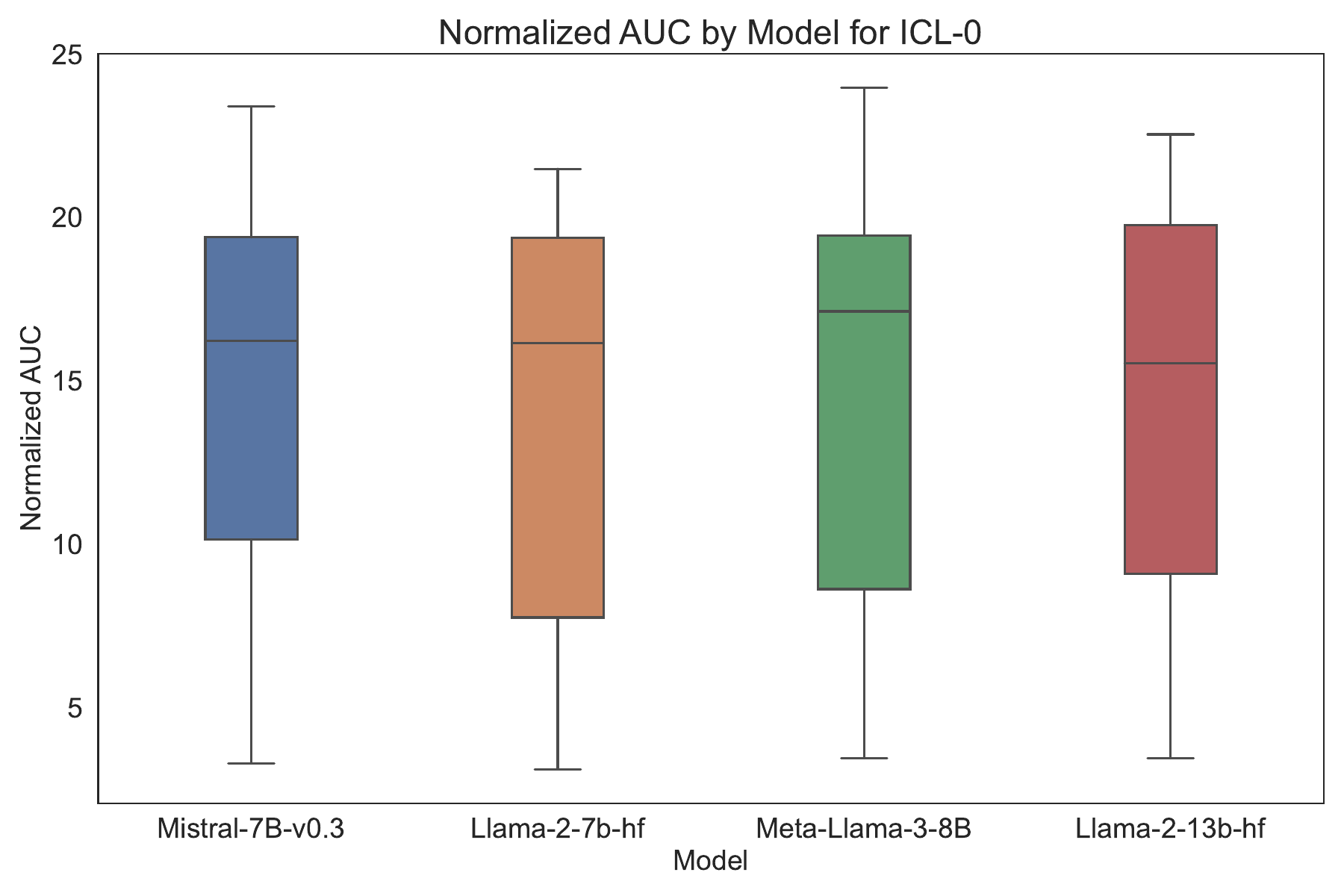}
    \caption{Normalized AUC by Model boxplot for ICL-0 experiments.}
    \label{fig:auc_boxplot_icl-0}
\end{figure}

\begin{figure}[H]
    \centering
    \includegraphics[width=\linewidth]{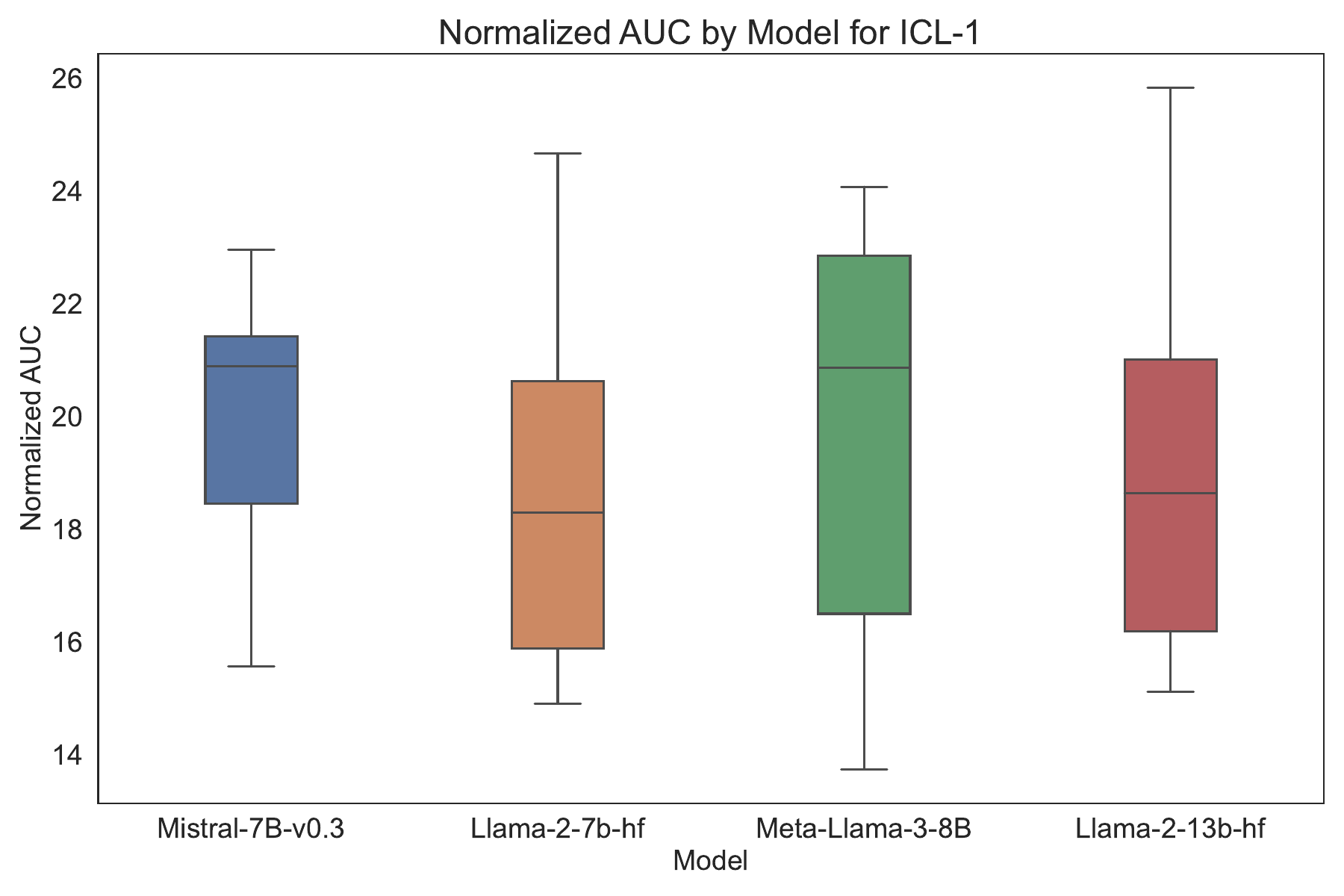}
    \caption{Normalized AUC by Model boxplot for ICL-1 experiments.}
    \label{fig:auc_boxplot_icl-1}
\end{figure}

\begin{figure}[H]
    \centering
    \includegraphics[width=\linewidth]{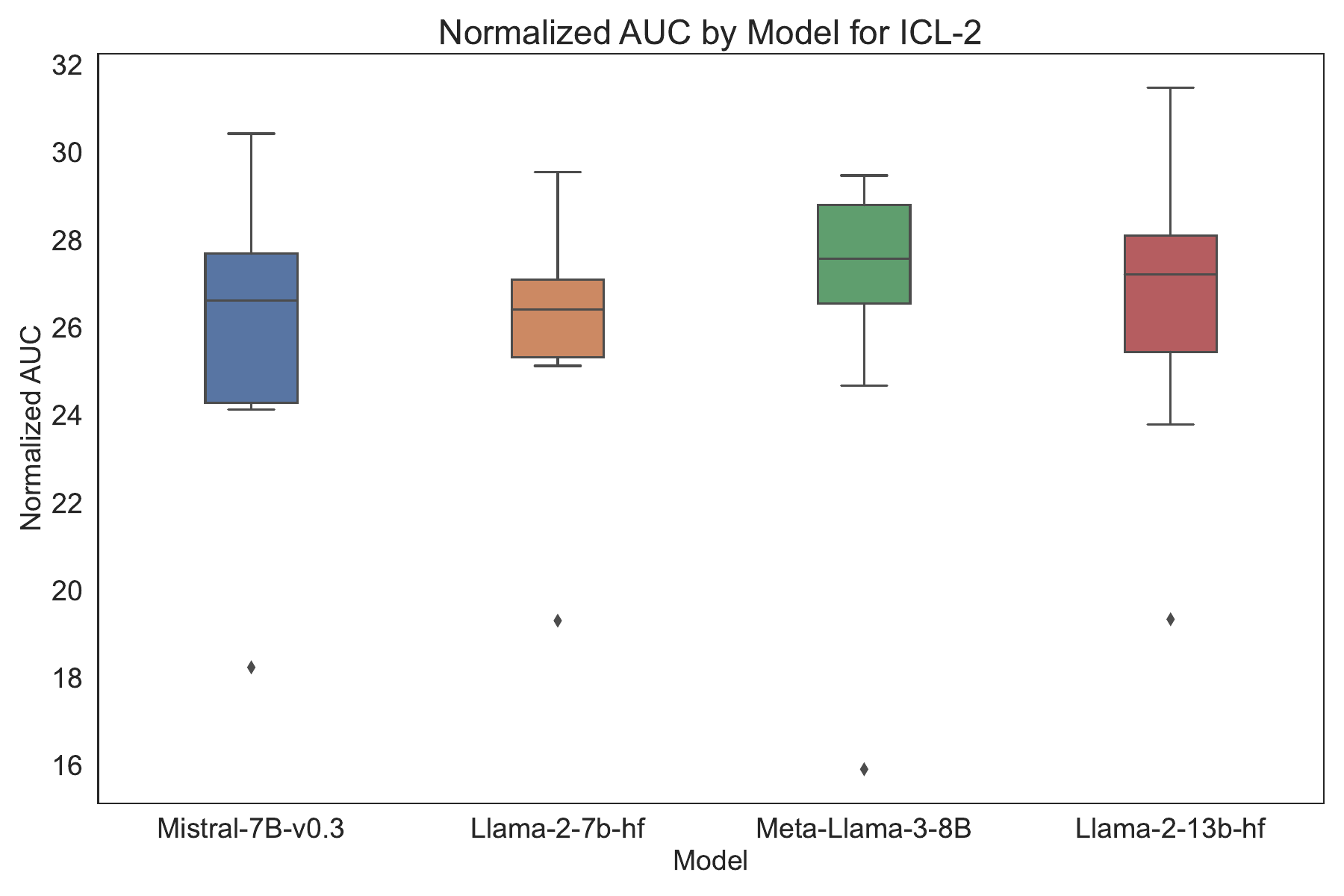}
    \caption{Normalized AUC by Model boxplot for ICL-2 experiments.}
    \label{fig:auc_boxplot_icl-2}
\end{figure}

\begin{figure}[H]
    \centering
    \includegraphics[width=\linewidth]{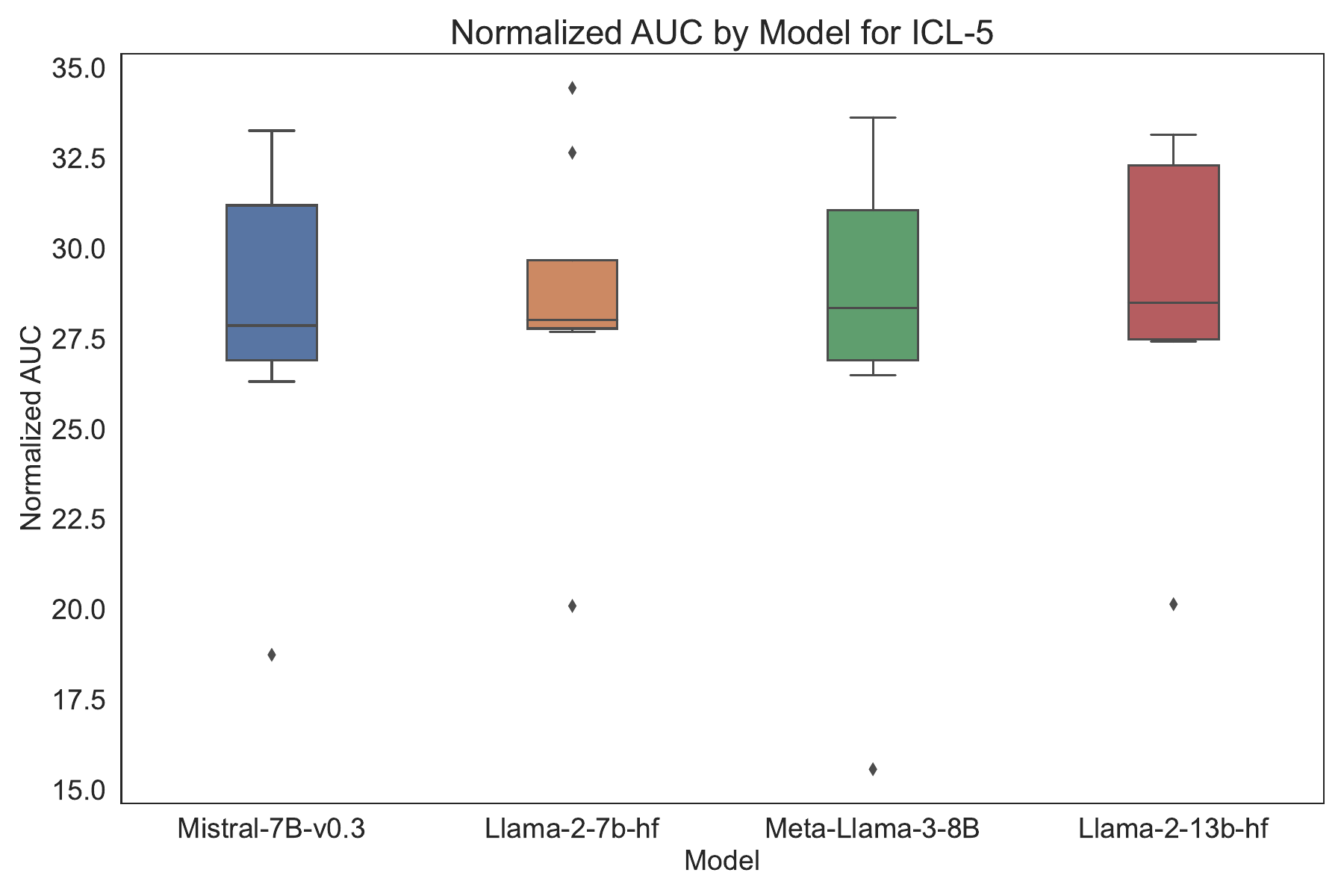}
    \caption{Normalized AUC by Model boxplot for ICL-5 experiments.}
    \label{fig:auc_boxplot_icl-5}
\end{figure}

\begin{figure}[H]
    \centering
    \includegraphics[width=\linewidth]{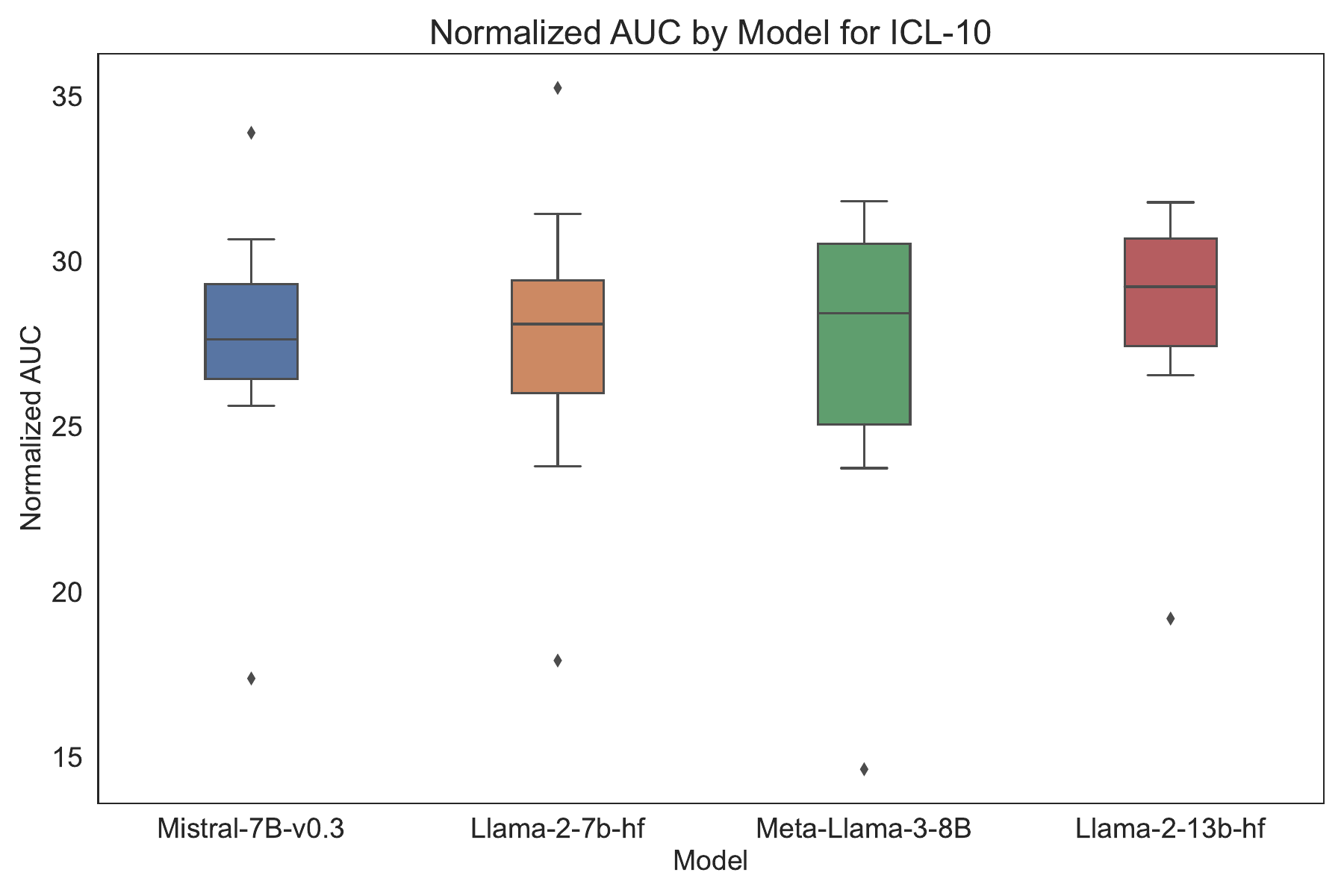}
    \caption{Normalized AUC by Model boxplot for ICL-10 experiments.}
    \label{fig:auc_boxplot_icl-10}
\end{figure}

\begin{figure}[H]
    \centering
    \includegraphics[width=\linewidth]{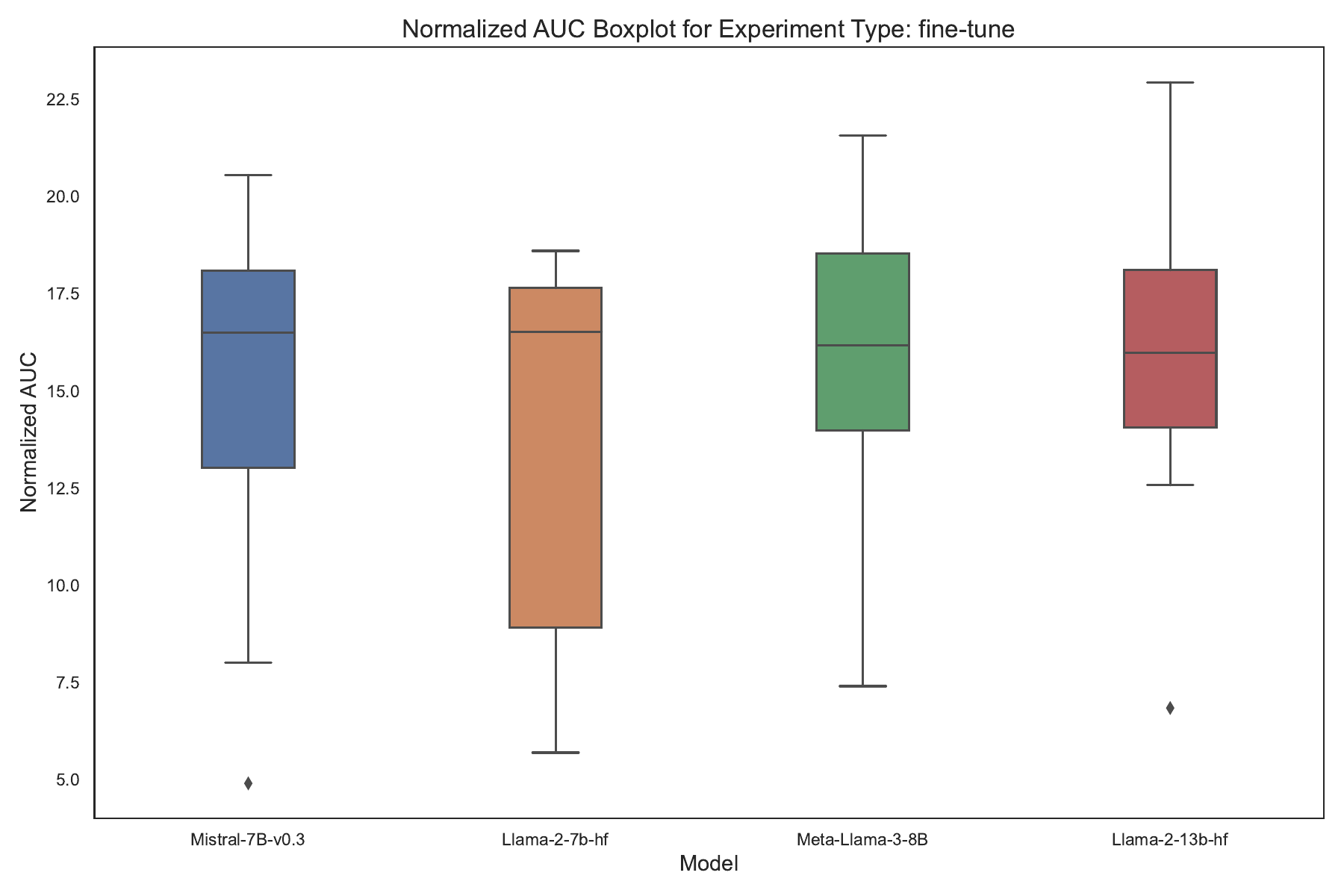}
    \caption{Normalized AUC by Model boxplot for SFT experiments.}
    \label{fig:5}
\end{figure}

% We observe that similar to the normalized AUC values presented in \ref{fig:auc_boxplot_icl}, the average normalized AUC values for different models are similar, although there are some outlier points with lower ID.

\section{Validating the TwoNN Estimator with the MLE Estimator}
\label{sec: validate-twonn}

\begin{figure}[H]
    \centering
    \includegraphics[width=\linewidth]{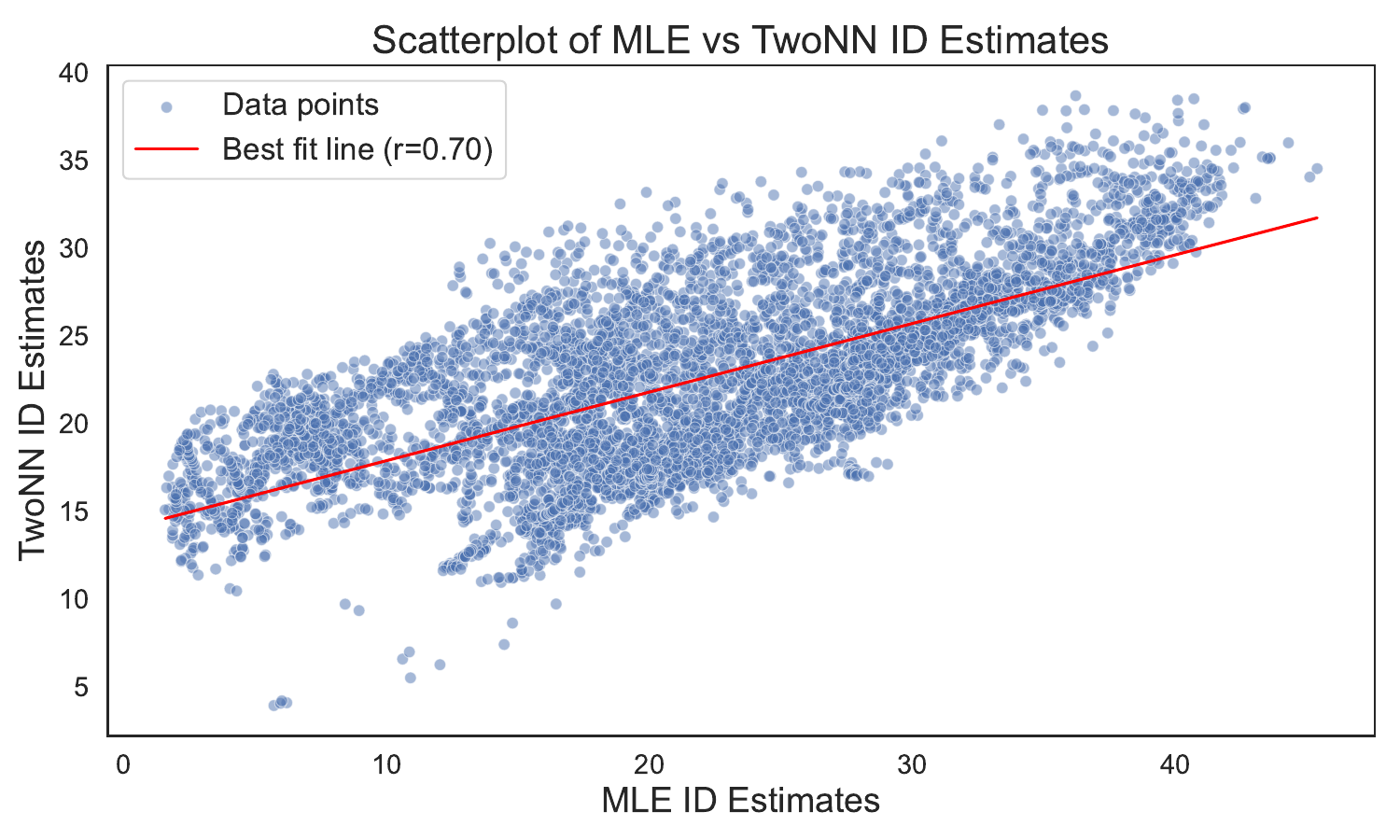}
    \caption{Scatterplot plotting ID estimation results for all experiments using the MLE and TwoNN Estimators.}
    \label{fig:validate_twonn}
\end{figure}

To assess the validity of our intrinsic dimension estimator, we calculate the intrinsic dimension for different combinations of (learning paradigm, dataset, model, layer) using the TwoNN estimator and Maximum Likelihood Estimator (MLE) introduced by \citet{levina2004maximum}. We use a neighborhood of size $k=50$ when applying MLE. We find that the estimates from the two estimators are correlated with $r=0.7$. While it is not possible to know the 'true' intrinsic dimensionality of the representations, high correlation between two separate estimators provides a sanity check for our choice of the TwoNN estimator.

\lstset{
    basicstyle=\ttfamily\footnotesize,
    breaklines=true,
    breakatwhitespace=true,
    postbreak=\mbox{\textcolor{red}{$\hookrightarrow$}\space}
}

\section{Dataset Generation Details}
\label{sec: dataset-curation-details}

\section{Dataset Details}

We include details about dataset generation below. We get prompts for all datasets except MMLU from the PromptSource library \cite{bach-etal-2022-promptsource}.

\subsection{QNLI}
Items for the training and validation splits in our QNLI experiments were taken from the official QNLI 'train' and 'validation' splits respectively.

\textbf{Prompt Template:}
\begin{lstlisting}
Does that sentence have all you need to answer the question "{{question}}"?
|||
{{answer_choices[label]}}
\end{lstlisting}

\textbf{Labels:} ['yes', 'no']

\subsection{CommonsenseQA}
Items for both the training and validation splits in our CommonsenseQA experiments were taken from the official CommonsenseQA 'train' split.

\textbf{Prompt Template:}
\begin{lstlisting}
Given the following options, what do you think is the correct answer to the question below:

{{question}}

Options:
{% for letter, t in zip(answer_choices, choices.text) %}
- {{letter}}: {{t}}
{% endfor %} |||
{{answerKey}}
{% endif %}
\end{lstlisting}

\textbf{Labels:} ['A', 'B', 'C', 'D']

\subsection{MMLU}
Items for both the training and validation splits in our MMLU experiments were taken from the official MMLU 'test' split.

\textbf{Prompt Template:}
\begin{lstlisting}
# generate input txt and output txt
letters = ['A', 'B', 'C', 'D']
choices = dataset_element['choices']

input_txt = f"{dataset_element['question']}\n\nA: {choices[0]}\nB: {choices[1]}\nC: {choices[2]}\nD: {choices[3]}\nAnswer:"

output_txt = letters[answer_idx]
combined = input_txt + output_txt
\end{lstlisting}

\subsection{SST-2}
Items for both the training and validation splits in our SST-2 experiments were taken from the official SST-2 'train' split.

\textbf{Prompt Template:}
\begin{lstlisting}
{{sentence}}
Question: Was that sentence {{"positive"}} or {{"negative"}}? Answer: ||| {{ answer_choices[label] }}
\end{lstlisting}

\textbf{Labels:} ['negative', 'positive']

\subsection{CoLA}
Items for both the training and validation splits in our CoLA experiments were taken from the official CoLA 'train' split.

\textbf{Prompt Template:}
\begin{lstlisting}
Does the following sentence make sense and use correct English? Please answer {{"yes"}} or {{"no"}}.
{{sentence}}
|||
{{ answer_choices[label] }}
\end{lstlisting}

\textbf{Labels:} ['no', 'yes']

\subsection{AGNews}
Items for the training and validation splits in our AGNews experiments were taken from the official AGNews 'train' and 'validation' splits respectively.

\textbf{Prompt Template:}
\begin{lstlisting}
What label best describes this news article?
{{text}} ||| 
{{answer_choices[label] }}
\end{lstlisting}

\textbf{Labels:} ['World politics', 'Sports', 'Business', 'Science and technology']

\subsection{MNLI}
Items for the training and validation splits in our MNLI experiments were taken from the official MNLI 'train' and 'validation\_matched' splits respectively.

\textbf{Prompt Template:}
\begin{lstlisting}
{{premise}} Are we justified in saying that "{{hypothesis}}"? Yes, no, or maybe? ||| {{ answer_choices[label] }}
\end{lstlisting}

\textbf{Labels:} ['Yes', 'Maybe', 'No']

\subsection{QQP}
Items for the training and validation splits in our QQP experiments were taken from the official QQP 'train' and 'validation' splits respectively.

\textbf{Prompt Template:}
\begin{lstlisting}
I'm an administrator on the website Quora. There are two posts, one that asks "{{question1}}" and another that asks "{{question2}}". I can merge questions if they are asking the same thing. Can I merge these two questions? ||| {{ answer_choices[label] }}
\end{lstlisting}

\textbf{Labels:} ['no', 'yes']

\end{document}